# A Comprehensive Survey of Text Classification Techniques and Their Research Applications: Observational and Experimental Insights

**Kamal Taha***
Department of Computer Science, Khalifa University, kamal.taha@ku.ac.ae

**Paul D. Yoo**
Department of Computer Science & Information Systems, University of London, Birkbeck College, UK, p.yoo@bbk.ac.uk

**Chan Yeun**
Center for Cyber-Physical Systems, Dept. of Electrical Engineering and Computer Science, Khalifa University, chan.yeun@ku.ac.ae

**Dirar Homouz**
Department of Physics, Khalifa University, dirar.homouz@ku.ac.ae

**Aya Taha**
Brighton College, Dubai, bc004814@brightoncollegedubai.ae

The exponential growth of textual data presents substantial challenges in management and analysis, notably due to high storage and processing costs. Text classification, a vital aspect of text mining, provides robust solutions by enabling efficient categorization and organization of text data. These techniques allow individuals, researchers, and businesses to derive meaningful patterns and insights from large volumes of text. This survey paper introduces a comprehensive taxonomy specifically designed for text classification based on research fields. The taxonomy is structured into hierarchical levels: research field-based category, research field-based sub-category, methodology-based technique, methodology sub-technique, and research field applications. We employ a dual evaluation approach: empirical and experimental. Empirically, we assess text classification techniques across four critical criteria. Experimentally, we compare and rank the methodology sub-techniques within the same methodology technique and within the same overall research field sub-category. This structured taxonomy, coupled with thorough evaluations, provides a detailed and nuanced understanding of text classification algorithms and their applications, empowering researchers to make informed decisions based on precise, field-specific insights.

**Keywords:** Text classification, Text data mining, Data science, Artificial Intelligence, deep learning.

## 1 Introduction

The advent of automated data acquisition tools and high throughput technologies has significantly contributed to the generation of vast volumes of text data, with sources like Wikipedia and Twitter leading the charge in this data explosion [1, 2]. This proliferation is further augmented by technologies such as cloud storage, sensor networks, and social networks, collectively adding to the vast data landscape [3]. However, the sheer volume and complexity of managing and processing such data can be daunting due to the extensive storage requirements and the high costs and time involved in processing. Text classification, as a critical component of text mining, has emerged as a pivotal technique in addressing these challenges [4].

Text classification specifically focuses on categorizing and organizing text data to facilitate easier management and analysis. These techniques leverage natural language processing, machine learning, and data mining to extract meaningful patterns and categorize text data, thereby transforming unstructured text into structured, actionable insights [5]. Applications of these methods are crucial in areas such as document classification, where texts are categorized based on their content, and information retrieval, which involves searching through large datasets to find relevant documents [2]. Additionally, text classification is pivotal in sentiment analysis, spam detection, and topic modeling, enabling better understanding and utilization of text data across various fields. The growing accessibility of powerful computing resources and open-source tools has significantly enhanced the feasibility of employing text mining techniques for both academic research and practical applications [7].

This paper aims to conduct a comprehensive evaluation of modern text classification algorithms through empirical and experimental assessments. We have developed a taxonomy system based on research fields that categorizes these algorithms into nested hierarchical levels, allowing for a more accurate and precise classification of techniques. Our analysis included a systematic review of papers that discuss specific text classification techniques, sourced from reputable publishers like IEEE and ACM. This ensures that our selected papers are up-to-date and reflect the current state-of-the-art in text classification [6]. Papers were ranked based on their novelty and relevance, with priority given to those providing comprehensive details on specific techniques.

Through this extensive evaluation, we aim to elucidate the strengths and weaknesses of various text classification techniques and their research fields. The insights provided will serve as a valuable resource for future research in this rapidly evolving field, offering a rigorous and nuanced understanding of modern algorithms and their practical applications.

* Corresponding author.

## 1.1 Motivation and Key Contributions

Existing survey papers on text classification often group algorithms into broad categories, which can lead to confusion. This broad categorization results in the use of the same metrics for evaluating disparate methods. For instance, algorithms may be grouped under the broad class of deep learning or a narrower category like CNN. However, comparing algorithms within these broad classes using the same metric or dataset can be inappropriate. A more specific classification is needed to address these challenges.

To overcome this issue, we propose a research field-based taxonomy that hierarchically classifies text classification algorithms into four layers, with each layer becoming more specific. We believe that comparing algorithms at the second and third layers of this taxonomy is more appropriate. Our taxonomy is structured into four tiers: research field-based category, research-based sub-category, methodology-based technique, methodology-based sub-technique, and applications of research fields utilized by the sub-technique/technique. This structure helps researchers better understand the relationships between different algorithms and the specific techniques they use.

In addition to proposing a detailed taxonomy, our study includes rigorous empirical evaluations. Moreover, we conduct experimental evaluations under the following five scenarios:

1. **Recurrent Neural Network-Based Data Science Classification**: Techniques for sentiment and sequence classification tasks, important for analyzing user-generated content, using the Yelp reviews dataset [9].
2. **Embedding-Based Data Science Classification**: Techniques for managing text with diverse formats and errors, crucial for news categorization, using the BBC News dataset [7].
3. **Pre-Trained-Based Classification Data Science**: Techniques for measuring performance in transfer learning for sentiment analysis, using the SST-2 dataset [8].
4. **Traditional Artificial Intelligence Techniques**: Evaluating their continued relevance and limitations in contemporary text classification tasks, using the Reuters-21578 dataset [10].
5. **Deep Learning Artificial Intelligence Techniques**: Handling large-scale, diverse text data for sentiment analysis and product review classification, supporting commercial applications, using Amazon product reviews dataset [11].

This methodology allows us to highlight and emphasize the subtle distinctions between closely related algorithms and techniques, aiding researchers in selecting the most appropriate methods for their specific tasks.

Through our research field-based taxonomy, empirical evaluations, and experimental comparisons, researchers can gain a nuanced and comprehensive understanding of text classification algorithms. This in-depth knowledge empowers them to make well-informed decisions, optimizing the impact and effectiveness of their research in the field of text data mining.

Figure 1 illustrates our research field-based taxonomy that arranges the algorithms for text classification into detailed classes using a hierarchical structure, consisting of research field-based category, research field-based sub-category, methodology technique, methodology sub-technique, and applications of research fields utilized by the sub-technique/technique.

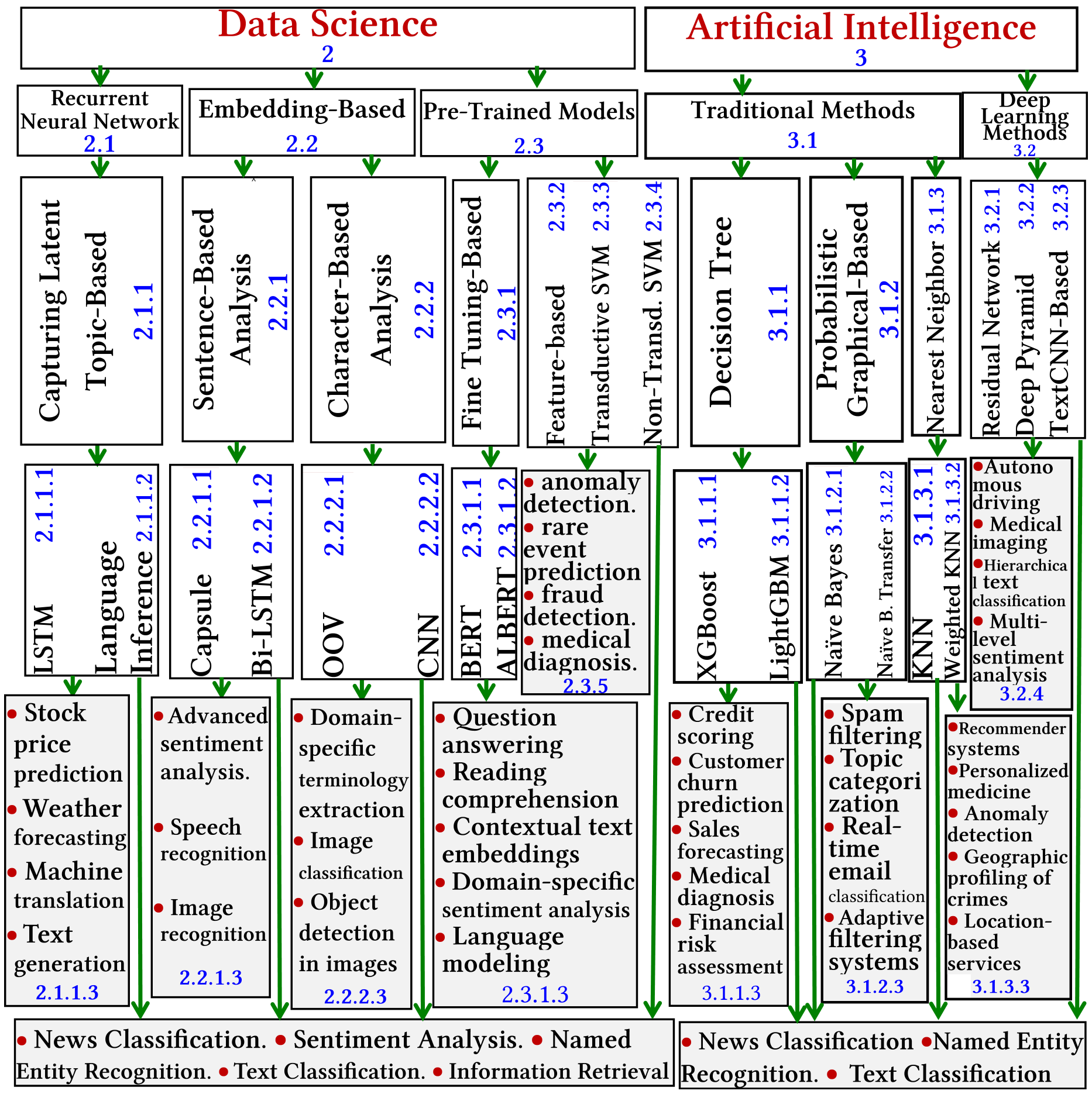

**Fig. 1:** Our research field-based taxonomy for text classification employs a detailed, hierarchical structure, consisting of the following levels: research field-based category → research field-based sub-category → methodology-based technique → methodology-based sub-technique. Each sub-technique or technique is linked to two boxes with a grey background that highlight its applications in research fields. One box contains unique applications specific to that sub-technique or technique, while the other box lists applications that can be utilized by all sub-techniques or techniques within the same research field-based category.

## 1.2 Current Survey Papers on the Topic

Liu et al. [1] classified topic analysis into five categories, including clustering, topic analysis, community discovery, contextual text mining, and text segmentation. Ignaczak et al. [2] categorized Neural Network Models into seven types, such as Convolutional Neural Network, Long Short-Term Memory, and Recurrent Neural Network. Joshi et al. [3] categorized text-based epidemic intelligence into six categories, including Topic model-based, Ontology enhanced, and Deep Learning-based. Li et al. [4] categorized DL models into ten types of classification, including News Classification, Dialog Act Classification, and Sentiment Analysis, among others. Tandel et al. [5] classified text mining techniques into three categories: Summarization, Clustering, and Information Extraction. Pen et al. [6] provided a thorough review of Deep Learning methods for Textual Emotion Analysis (TEA), classifying existing TEA approaches based on text structures and linguistic types. The authors identified four categories, including text-oriented monolingual methods, text conversations-oriented monolingual methods, cross-linguistic methods, and emoji-oriented methods. Minaee et al. [12] reviewed deep learning models for text classification tasks like sentiment analysis, news categorization, topic classification, question answering, and natural language inference, categorizing them by architecture: RNNs, CNNs, attention mechanisms, Transformers, and Capsule Nets.

Francia et al. [13] conducted a literature review of ML- and DL-based techniques for predicting rule decisions, examining the text pre-presentation techniques used to transform legal text into suitable input formats for learning algorithms. Sarwar et al. [14] proposed a taxonomy for neural-network embedding techniques, categorizing them into: Embedding from Language Models and Word2Vec. Pham et al. [15] surveyed reviews recent deep learning-based approaches, particularly TG-GNN (text graph neural network) techniques, for text representation learning and classification, highlighting their capabilities, advantages, and limitations. They concluded with comparative studies on TG-GNN models and discusses future research directions to address existing challenges in the field.

Li et al. [16] surveyed text classification methods from 1961 to 2021, categorizing them into traditional and deep learning-based techniques. They discussed technical developments, benchmark datasets, evaluation metrics, and compared various techniques. The paper concludes with insights on key implications, future research directions, and challenges in the field. It reviews performance improvements of traditional and deep learning models, datasets, evaluation metrics, and presents quantitative results of leading models, outlining future research challenges. Gasparetto et al. [17] reviewed recent models, focusing on the data flow from raw text to output labels, and highlighted differences between traditional and deep learning methods. They also provided an overview of English language datasets, introduces two new multilabel datasets, presents experimental results, and discusses future research challenges related to the robustness of deep learning.

# 2 Data Science Research Field-Based Category

Data science for text classification utilizes algorithms and computational techniques to organize and categorize large sets of textual data into predefined categories. This process involves natural language processing (NLP) methods to transform raw text into structured data, which can then be analyzed and classified using machine learning models. Applications of text classification include sentiment analysis, spam detection, and topic labeling, enhancing the ability to efficiently manage and interpret vast amounts of text information.

## 2.1 Recurrent Neural Network-Based Classification Sub-Category

Recurrent Neural Network (RNN)-based language models incorporate prior information by analyzing the placement of words. To start, each input word is converted into a unique vector via word embedding technology. These vectors are then sequentially processed through RNN cells. The RNN cells' output vectors are of the same dimension as the input vectors and are forwarded to the next hidden layer.

### 2.1.1 Capturing Latent Topic Technique

#### *2.1.1.1 Long Short-Term Memory-Based Classification Sub-Technique*

This technique utilizes RNNs to reveal inherent relationships within a given sequential data. This approach can effectively capture the sequence of events in predictive tasks while retaining information for an extended period. It is adept at handling variable-length sequences.

*The rationale behind the usage of the technique:* To enhance a model's ability to learn grammatical dependencies, it is essential to integrate an additional module into a neural network that can recognize the information that will be needed at a later point in a sequence, as well as the moment when it becomes irrelevant.

*Conditions for the optimal performance of the technique:* The performance of this technique can be enhanced through supervised training on a set of training sequences, utilizing an optimization algorithm such as gradient descent, and backpropagation through time to calculate gradients.

*Limitations of the technique:* To acquire proficiency, the technique requires supplementary training data. It may have restricted effectiveness in classification tasks if the input data is not presented in a sequential format, rendering it unsuitable for all types of data.

**Table 1:** Featuring and evaluating research papers that have employed Long Short-Term Memory-based techniques.

| Paper/ Year | Dataset | Scalability | Interpretability | Accuracy | Efficiency | Description |
|---|---|---|---|---|---|---|
| [18] 2021 | Author Collected | Good | Unsatisfactory | Good | Unsatisfactory | The authors proposed a three-layer Bi-LSTM for detecting sarcasm |
| [19] 2020 | SogouCS, Toutiao | Acceptable | Unsatisfactory | Good | Fair | The authors used Bi-LSTM and hierarchical attention for categorizing Chinese text |
| [20] 2021 | Myers Briggs Type Indicator | Fair | Unsatisfactory | Good | Unsatisfactory | The authors merged LSTM and CNN to distinguish personality traits from text content |
| [21] 2020 | IMDB, Amazon Alexa Reviews | Good | Fair | Acceptable | Acceptable | The authors introduced a GCN-based approach using BiLSTM method to handle contextual dependencies |
| [22] 2020 | Author Collected | Acceptable | Acceptable | Acceptable | Unsatisfactory | The authors recommended Bi-LSTM for assigning emotion labels to psychiatric texts |

#### 2.1.1.2 *Language Inference-Based Classification Sub-Technique*

The technique employs rule-based systems, linear classifiers, or neural networks to discover correlations and extract semantic features. It assesses whether a statement can be deduced from a given text premise by categorizing it into one of three groups.

*The rationale behind the usage of the technique:* Considering established facts and previous knowledge is crucial for classification. NLI can detect correlations between contexts. It acquires semantic characteristics. By incorporating syllogistic reasoning into preexisting facts, NLI can generate a novel dataset of knowledge. Syllogistic reasoning generates new knowledge by combining preexisting facts.

*Conditions for the optimal performance of the technique:* By masking all words within a unit during word representation training instead of just one, the performance of the technique can be enhanced by facilitating the implicit acquisition of prior knowledge. This method enables the learning of longer semantic dependencies that aid in guiding the process of word embedding learning.

*Limitations of the technique:* Obtaining a complete set of logical and commonsense inferences from a training corpus can pose a difficulty, especially in modern data-intensive tasks, as the corpus is often too small. As a result, the technique may struggle to generalize beyond the specific examples provided in the training corpus, leading to suboptimal performance on data.

**Table 2:** Featuring and evaluating research papers that have employed Recurrent Neural Network-based techniques.

| Paper/ Year | Dataset | Scalability | Interpretability | Accuracy | Efficiency | Description |
|---|---|---|---|---|---|---|
| [23] 2019 | NLPCC-DBQA, MSRA-NER | Good | Unsatisfactory | Good | Fair | The authors integrated knowledge into pre-trained language models using language inference |
| [24] 2021 | COVID-19 news dataset | Acceptable | Fair | Fair | Fair | The authors proposed using Gumbel-Softmax to model topic-document inference, sampling topic proportions in an autoencoder |
| [25] 2017 | Stanford Natural Language Inference | Acceptable | Unsatisfactory | Good | Unsatisfactory | The authors acquired universal sentence representations via NLI data |

#### 2.1.1.3 *Applications of Research Fields Utilizing LSTM and Language Inference Sub-Techniques*

- **Stock price prediction**: LSTMs can capture temporal dependencies in historical stock prices, allowing them to predict future prices by analyzing sequential data patterns. Language inference can enhance these predictions by incorporating sentiment analysis of relevant financial news.
- **Weather forecasting**: LSTMs can model and predict weather patterns by learning from historical weather data, effectively capturing long-term dependencies. Language inference can integrate weather reports and social media data to improve forecasting accuracy.
- **Machine translation**: LSTMs can be used in sequence-to-sequence models to translate text from one language to another by understanding and generating the appropriate sequence of words. Language inference helps in maintaining context and semantic meaning during translation.
- **Text generation**: LSTMs can generate coherent and contextually relevant text by learning from large text corpora, predicting the next word in a sequence. Language inference ensures the generated text aligns with the intended meaning and context.
- **News classification**: LSTMs can classify news articles by learning temporal patterns in the text, identifying topics and categories. Language inference aids in understanding nuanced meanings and categorizing articles accurately.
- **Sentiment analysis**: LSTMs can analyze sequences of words in text to determine sentiment by capturing dependencies and context. Language inference helps in accurately identifying sentiments, even in complex sentences.
- **Named Entity Recognition (NER)**: LSTMs can recognize and classify named entities in text by learning patterns in sequential data. Language inference improves the identification of entities by considering context and semantic relationships.
- **Text classification**: LSTMs can classify text into predefined categories by learning from sequences of words and their contextual dependencies. Language inference enhances classification by understanding the deeper meaning of the text.
- **Information retrieval**: LSTMs can improve information retrieval by understanding query context and matching it with relevant documents. Language inference refines search results by interpreting the intent and meaning behind queries.

## 2.2 Embedding-Based Classification Sub-Category

Embedding techniques can effectively capture connections between linguistic units at different levels. Character-level embedding represents characters as vectors in a high-dimensional space, allowing the model to capture their structural and phonetic characteristics. Word-level embedding represents words as dense vectors in a high-dimensional space, helping the model associate similar words and comprehend a word's meaning based on its use in a sentence. Sentence-level embedding represents sentences as vectors in a high-dimensional space, allowing the model to capture overall meaning by considering individual word relationships and contexts.

### 2.2.1 Sentence-Based Technique

#### 2.2.1.1 *Capsule-based Sub-Technique*

In this technique, dynamic routing involves capsules (groups of neurons) in a layer capturing feature connections. The coupling coefficients are iteratively updated to transfer capsules from lower layers to higher ones, consolidating their transformations to establish relationships.

*The rationale behind the usage of the technique:* In a capsule-based structure, the routing procedure captures part-whole relationships and collects additional information iteratively to represent a sentence. By regulating self-loops, it extends the multi-hop attention technique, enabling multi-step attention within a single layer.

*Conditions for the optimal performance of the technique:* The addition of an attention mechanism to the capsule structure can enhance the text classification performance of this technique. Integrating syntactic information can also mitigate the problem of focusing on words that are not syntactically related, whether they are close or distant.

*Limitations of the technique:* During sentiment inference, the capsule network is unable to concentrate on aspect-specific words. The dynamic routing mechanism operates independently of backpropagation, resulting in slower performance and reduced efficacy. Additionally, it may mistakenly focus on words that are not syntactically related.

**Table 3: Featuring and evaluating research papers that have employed Capsule-based techniques.**

| Paper/ Year | Dataset | Scalability | Interpretability | Accuracy | Efficiency | Description |
|---|---|---|---|---|---|---|
| [26] 2019 | Yelp, Amazon, Twitter | Unsatisfactory | Acceptable | Good | Unsatisfactory | They introduced an OOV-based method for encapsulating sentence-level semantic representations into semantic capsules, which transfer knowledge at the document level to improve semantic understanding of sentences |
| [27] 2020 | Twitter, Lap14, Rest14 | Fair | Good | Good | Fair | They used syntactic knowledge and combined a knowledge-guided capsule attention network with a Bi-LSTM network to achieve better results |
| [28] 2018 | Movie Review, Stanford Sentiment | Unsatisfactory | Acceptable | Good | Fair | They proposed an attention capsule structure for aspect-based sentiment analysis, where aspect term embeddings replac traditional capsule queries |
| [29] 2017 | Political Blog, DBLP | Unsatisfactory | Fair | Acceptable | Acceptable | They presented a framework for attributed graph embedding that clusters content-enriched graphs and encodes them into low-dimensional representations. |

### 2.2.1.2 Bidirectional LSTM-based Sub-Technique

The technique employs a Bidirectional LSTM, which is a sequence processing method that leverages two LSTMs, with one processing input in the forward direction and the other processing it in the backward direction, to encode input sentences.

*The rationale behind the usage of the technique:* With the utilization of BiLSTMs, the network can access a larger volume of information, thereby improving the algorithm's contextual awareness. This allows it to comprehend not only the word in question but also the words preceding and succeeding it in a sentence.

*Conditions for the optimal performance of the technique:* Integrating 2D convolution into BiLSTM can enhance contextual comprehension and capture features, resulting in improved text classification. The addition of CNN layers to BiLSTM can address aspect category clustering. 2D convolutional layers are effective at detecting local patterns and features within data.

*Limitations of the technique:* Compared to LSTM, this technique is considerably slower and requires more time for training. This can be attributed to the increased complexity and computational demands associated with the incorporation of 2D convolutional layers. The inclusion of these layers can lead to longer training times and slower performance during inference.

**Table 4:** Featuring and evaluating research papers that have employed Bidirectional LSTM-based techniques.

| Paper/ Year | Dataset | Scalability | Interpretability | Accuracy | Efficiency | Description |
|---|---|---|---|---|---|---|
| [30] 2016 | Twitter | Fair | Acceptable | Good | Unsatisfactory | The authors utilized a BiLSTM network with word embeddings of 300 dimensions and a learning rate of 0.001 to analyze sentiment |
| [31] 2017 | SemEval 2014 | Unsatisfactory | Fair | Good | Fair | The authors introduced a technique involves the use of two BiLSTMs to represent context and target in text |
| [32] 2017 | SemEval workshops | Unsatisfactory | Fair | Acceptable | Unsatisfactory | The authors implemented CNN layers and a BiLSTM network to extract aspect terms and classify aspect categories |
| [33] 2022 | AMZN, AAPL, NFLX | Fair | Acceptable | Acceptable | Fair | The authors used a bidirectional LSTM network with attention to predict stock prices, reducing the number of parameters significantly |

### 2.2.1.3 Applications of Research Fields Utilizing Capsule-Based and Bi-LSTM Sub-Techniques

- **Advanced sentiment analysis**: Capsule networks can capture spatial hierarchies and relationships between features in text, enhancing sentiment detection. Bi-LSTMs can process text bidirectionally, understanding context from both past and future words to accurately determine sentiment.
- **Speech recognition**: Capsule networks can model hierarchical relationships in audio features, improving the robustness of speech recognition. Bi-LSTMs can capture temporal dependencies in speech sequences from both directions, enhancing the accuracy of transcriptions.
- **Image recognition**: Capsule networks excel in recognizing spatial relationships and pose variations in images, providing superior accuracy in image recognition tasks. Bi-LSTMs, although primarily used for sequential data, can process sequences of image patches or features for temporal-based image analysis.
- **News classification**: Capsule networks can identify complex feature hierarchies in news articles, improving classification accuracy. Bi-LSTMs can leverage bidirectional context to better understand and classify news content based on sequential word patterns.
- **Sentiment analysis**: Capsule networks enhance sentiment analysis by capturing intricate relationships between features in text. Bi-LSTMs provide a comprehensive understanding of sentiment by processing text from both directions, capturing nuanced context.

- **Named Entity Recognition (NER)**: Capsule networks can effectively model relationships between entities and their contexts in text. Bi-LSTMs enhance NER by considering bidirectional dependencies, improving entity identification and classification accuracy.
- **Text classification**: Capsule networks improve text classification by capturing hierarchical dependencies and relationships within text. Bi-LSTMs enhance classification by utilizing bidirectional context, providing a deeper understanding of the text.
- **Information retrieval**: Capsule networks can capture complex feature relationships, improving the retrieval of relevant information. Bi-LSTMs enhance retrieval by understanding bidirectional context in queries and documents, refining search results.

### 2.2.2 Character-Level-Based Technique

#### *2.2.2.1 Out-Of-Vocabulary-based Sub-Technique*

This technique utilizes language modeling to infer the meaning of an OOV word in a sentence by comparing it with similar sentences. This involves these steps: tokenization, training, and saving a model based on a training corpus.

*The rationale behind the usage of the technique:* The utilization of language modeling to forecast the meaning of an OOV word leads to consistent performance for unfamiliar words in both intrinsic and extrinsic tasks. Smoothing techniques are employed to avoid zero-probability occurrences, and the OOV word can be replaced with a synonym as an alternative solution.

*Conditions for the optimal performance of the technique:* A Bi-directional RNN is capable of predicting embeddings for OOV words based on their context. The use of unknown (UNK) to replace unknown words reduces the reliance on the training corpus, thus generating dependable embeddings for OOV words. This can aid in entity recognition tasks by providing vector representations.

*Limitations of the technique:* The limitations include its reliance on the training corpus for word embeddings, the difficulty of recognizing OOV words that have not been seen before, the inability to predefine the vocabulary for average-sized domains, and the exacerbation of the OOV problem when dealing with real-world datasets containing slang and typos that were not included in the training

**Table 5:** Featuring and evaluating research papers that have employed Out-Of-Vocabulary techniques.

| Paper/ Year | Dataset | Scalability | Interpretability | Accuracy | Efficiency | Description |
|---|---|---|---|---|---|---|
| [34] 2018 | Synthetic dataset | Unsatisfactory | Acceptable | Good | Acceptable | The authors proposed a deep CNN that utilizes character-level embeddings for OOV words to classify multi-lingual texts. |
| [35] 2022 | Tipster, Robust | Unsatisfactory | Fair | Acceptable | Fair | They expanded queries using a neural generative model that employs the GPT-2 to generate multiple texts and estimate word importance based on frequency. |

#### *2.2.2.2 CNN-Based Sub-Technique*

To generate a distributed representation based on words, this technique takes a sequence of encoded characters as input. At the word level, character-level features are extracted to construct this representation.

*The rationale behind the usage of the technique:* The generation of a distributed representation from words can be applied to both distributed and discrete word embeddings, irrespective of whether one has knowledge of the syntactic or semantic structures. This enables the capturing of semantic features from the data.

*Conditions for the optimal performance:* The integration of deep contextualized word representations such as ELMo and BERT into this technique can enhance the feature representations that are not reliant on the decoder, resulting in improvement in performance.

*Limitations of the technique:* The optimal functioning of this technique usually necessitates the use of large datasets. Its efficacy is contingent on various factors, including the size of the dataset, text curation, and the selection of the alphabet.

**Table 6:** Featuring and evaluating research papers that have employed CNN-based techniques.

| Paper/ Year | Dataset | Scalability | Interpretability | Accuracy | Efficiency | Description |
|---|---|---|---|---|---|---|
| [36] 2015 | AG's News, DBPedia | Acceptable | Fair | Good | Acceptable | The authors introduced a new approach that employs a character-level CNN framework that extracts semantic information from character-level signals. |
| [37] 2021 | CTB5, CTB6, CTB7 | Acceptable | Unsatisfactory | Fair | Good | They investigated character-level dependency parsing in Chinese and utilized deep CNNs. They integrated a CNN-based biaffine parser into a Chinese BERT model to improve encoding, resulting in better performance for POS tagging. |
| [38] 2018 | Academia, SNS | Fair | Unsatisfactory | Good | Good | The authors proposed a graph neural network framework that combines structural and attribute embedding for predicting alignment labels. It considers both structural and attribute aspects |

#### *2.2.2.3 Applications of Research Fields Utilizing OOV and CNN Sub-Techniques*

- **Domain-specific Terminology Extraction:** OOV techniques handle unknown or rare terms by mapping them to similar known terms or using context to infer their meaning, while CNNs can capture local patterns and hierarchies in text to identify and extract domain-specific terms accurately.
- **Image Classification:** CNNs excel in image classification by learning spatial hierarchies of features through convolutional layers, enabling them to classify images into predefined categories with high accuracy.

- **Object Detection in Images:** CNNs perform object detection by applying convolutional filters to identify objects within images, using techniques like region proposals and bounding box regressions to locate and classify multiple objects within a single image.
- **News Classification:** CNNs classify news articles by learning hierarchical features from text data, while OOV techniques manage out-of-vocabulary words to ensure accurate classification even with rare or unseen terms.
- **Sentiment Analysis:** CNNs analyze text for sentiment by detecting patterns and hierarchical structures in sentences, and OOV techniques handle unknown words to ensure sentiment is accurately captured despite the presence of rare terms.
- **Named Entity Recognition:** CNNs can be used to identify named entities in text by recognizing patterns and sequences associated with names, places, and organizations, while OOV techniques manage rare or new entities to improve recognition accuracy.
- **Text Classification:** CNNs classify text by learning and extracting relevant features from the input text, and OOV techniques ensure that even uncommon words are appropriately considered during classification.
- **Information Retrieval:** CNNs enhance information retrieval by learning relevant features and patterns in queries and documents, while OOV techniques handle rare terms to improve the matching and relevance of retrieved information.

## 2.3 Pre-Trained-Based Classification Sub-Category

This technique uses pre-trained language models as the starting point for building a new text classification model. The pre-trained language model is used as a feature extractor, and the output of the model is fed into a new classification model. The pre-trained language model is typically used to extract contextualized embeddings, which are then used to train a new model to classify text data. During fine-tuning, the pre-trained language model is trained on the specific task by updating the weights of the model

### 2.3.1 Fine-Tuning-Based Classification Technique

This technique takes a pre-trained language model and trains it on a classification task using a small amount of task-specific labeled data. During fine-tuning, the pre-trained model is adjusted to fit the classification task by updating the model's weights based on labeled data.

#### *2.3.1.1 BERT-Based Technique Sub-Technique*

In this technique, a multi-layer transformer encoder is employed to process data. The encoder extracts meaningful representations from the input data. The input tokens are embedded using a token embedding layer. This layer maps each input token to a vector representation in a high-dimensional space, which is passed through the transformer encoder.

*The rationale behind the usage of the technique:* By processing the input text bidirectionally (i.e., considering both the preceding and following words and the surrounding context of each word), a model can capture a greater amount of contextual information. This is crucial for accurately classifying text, as it allows the model to infer the meaning of the words based on their context and predict the label of the current word.

*Conditions for the optimal performance of the technique:* The technique can be enhanced by using a random sequence to predict training tokens instead of relying solely on BERT's masked language model, which only predicts masked tokens. Additionally, performance can be improved by sharing parameters between layers.

*Limitations of the technique:* (1) its large size, attributable to its training structure and corpus, (2) slow training time resulting from its size and numerous weights that need to be updated, and (3) increased computational time.

**Table 7:** Featuring and evaluating research papers that have employed BERT-based techniques.

| Paper/ Year | Dataset | Scalability | Interpretability | Accuracy | Efficiency | Description |
|---|---|---|---|---|---|---|
| [39] 2019 | GLUE, SQuAD | Good | Unsatisfactory | Good | Unsatisfactory | The authors used fine-tuning a pre-trained BERT model (768 dim) to improve ATSA with a learning rate of 2e-5 |
| [40] 2019 | RRC dataset | Good | Unsatisfactory | Acceptable | Unsatisfactory | The authors improved ATSA by post-training BERT (768 dim) with a maximum length of 320. |
| [41] 2019 | Laptop, Twitter | Fair | Unsatisfactory | Acceptable | Acceptable | The authors used BERT's hyper-parameter settings for fine-tuning. |
| [42] 2019 | SemEval 2014 | Acceptable | Unsatisfactory | Fair | Fair | The authors introduced an attentional encoder network to achieve target-specific learning with BERT |

#### *2.3.1.2 ALBERT-Based Classification Sub-Technique*

ALBERT is based on the same architecture as BERT but uses parameter reduction techniques to make the model more computationally efficient. It achieves this by sharing parameters across layers and using cross-layer parameter sharing to reduce the number of model parameters. Parameters from a layer are shared with a layer that is not adjacent to it.

*The rationale behind the usage of the technique:* The issue of computationally expensive training and running of language models, such as BERT, can be addressed by implementing two parameter simplification techniques - parameter sharing and factorization. This results in a reduction of parameters without sacrificing the model's performance.

*Conditions for the optimal performance of the technique:* ALBERT, which utilizes absolute position embeddings, can achieve better performance by right-padding input sequences rather than left-padding them. The reason behind this is that right-padding places padding tokens at the end of the sequence, enabling the model to distinguish between input tokens and padding tokens.

*Limitations of the technique:* ALBERT suffers the potential for bias, which may stem from fine-tuning and other training procedures. If the fine-tuning process relies on partial data or employs prejudiced evaluation metrics, the resulting model may inherit the biases. If the training data is biased, the model may acquire a bias towards making predictions based on that data.

**Table 8:** Featuring and evaluating research papers that have employed ALBERT-based techniques.

| Paper/ Year | Dataset | Scalability | Interpretability | Accuracy | Efficiency | Description |
|---|---|---|---|---|---|---|
| [43] 2020 | SQuAD2.0, NewsQA | Good | Unsatisfactory | Good | Acceptable | The authors introduced a pre-trained ALBERT-based technique for understanding Korean |
| [44] 2021 | Restaurant &Laptop datasets | Acceptable | Unsatisfactory | Good | Fair | They developed a method to enhance sentence representation with interactive information about aspects using ALBERT to connect sentences. |
| [45] 2022 | Twitter, Weibo | Fair | Unsatisfactory | Acceptable | Fair | The authors created an ALBERT-based model to encode audio information. |

#### 2.3.1.3 *Applications of Research Fields Utilizing BERT and ALBERT Sub-Techniques*

- **Question answering**: BERT and ALBERT utilize bidirectional transformers to understand the context of a question and its corresponding text, providing accurate and relevant answers by capturing the nuances of both. ALBERT, with its parameter-sharing mechanism, improves efficiency and scalability while maintaining accuracy.
- **Reading comprehension**: BERT and ALBERT excel at reading comprehension tasks by deeply understanding the context and relationships within the text, enabling them to answer questions based on passages with high accuracy. ALBERT's reduced model size enhances performance without sacrificing comprehension capabilities.
- **Contextual text embeddings**: BERT and ALBERT generate rich contextual embeddings by processing text bidirectionally, capturing the meaning of words in context. These embeddings can be used to enhance various NLP tasks by providing a deeper understanding of text semantics.
- **Domain-specific sentiment analysis**: BERT and ALBERT can be fine-tuned on domain-specific datasets to capture the unique sentiment expressions within a particular field, improving sentiment analysis accuracy. ALBERT's efficiency allows for scalable fine-tuning on large domain-specific corpora.
- **Language modeling**: BERT and ALBERT can be used for language modeling by predicting masked words in a sentence, learning bidirectional context. ALBERT's efficient design makes it suitable for large-scale language modeling tasks.
- **News classification**: BERT and ALBERT can classify news articles by understanding the context and relationships within the text, leading to accurate categorization. ALBERT's efficiency enables faster processing of large volumes of news data.
- **Sentiment analysis**: BERT and ALBERT perform sentiment analysis by leveraging their deep contextual understanding of text, capturing subtle nuances in sentiment expressions. ALBERT's compact architecture ensures efficient sentiment analysis without compromising accuracy.
- **Named Entity Recognition (NER)**: BERT and ALBERT excel at NER by understanding the context and relationships within text, accurately identifying and classifying named entities. ALBERT's efficiency allows for scalable NER applications.
- **Text classification**: BERT and ALBERT classify text by capturing the contextual meaning and relationships within the text, leading to high classification accuracy. ALBERT's parameter-sharing mechanism ensures efficient and scalable text classification.
- **Information retrieval**: BERT and ALBERT improve information retrieval by understanding the context of queries and documents, providing more relevant search results. ALBERT's efficiency enhances the speed and scalability of information retrieval systems.

### 2.3.2 Feature Embedding from Language Classification Technique

Feature embedding is the process of representing text data as numerical features that encapsulate the pertinent information within the text. This technique involves using a bidirectional LSTM model to generate word vectors, taking into account the context in which a word appears. By processing the word and its surrounding context in both directions, the LSTM leverages its internal states to accurately capture the word's meaning within its specific context. These numerical features are then utilized in classification algorithms, influencing their performance. The choice of feature embedding method is crucial for the algorithms' effectiveness.

*The rationale behind the usage of the technique:* By representing the embedding of a word based on the entire sentence that contains that word, the embeddings can capture the meaning of the word in that specific context. This allows the generation of embeddings for the same word used in various contexts across different sentences.

*Conditions for the optimal performance of the technique:* To improve the technique incorporate a bi-LSTM model, a variant of LSTM model. LSTM is a type of RNN that can capture the long-term dependencies and sequential patterns in a sequence of inputs.

*Limitations of the technique:* Words can have multiple meanings depending on the context in which they are used, and the technique may struggle to accurately identify the intended meaning in every instance (word meanings in distinct sentences rely on contextual subtleties). The technique may not efficiently handle unfamiliar words.

**Table 9:** Featuring and evaluating research papers that have employed Feature Embedding from Language techniques.

| Paper/ Year | Dataset | Scalability | Interpretability | Accuracy | Efficiency | Description |
|---|---|---|---|---|---|---|
| [46] 2017 | Newsgroup dataset, Reuter dataset | Acceptable | Fair | Good | Good | The authors introduces a semantic word processing technique for text categorization using the Back Propagation Lion (BP Lion) algorithm to improve neuron weight updates. Experiments on the 20 Newsgroup and Reuter datasets demonstrate that the proposed method reduces dimensionality and enhances classification accuracy. |
| [47] 2018 | 20 Newsgroup, Reuter, WebKB, RCV1 | Good | Fair | Good | Good | The authors proposed an incremental learning technique for text classification using a contextual-semantic word processing and hybrid fuzzy neural network for text classification that adapts dynamically to new data. The method, incorporating the BPLion Neural Network with fuzzy bounding and Lion Algorithm (LA), outperforms existing techniques (I-BP, FI-BP, I-BPLion) in accuracy. |
| [48] 2015 | Flickr , Youtube | Good | Unsatisfactory | Good | Good | The authors developed a network embedding framework for enhancing both global and local objective functions in various information networks |
| [49] 2020 | AGNews, Yelp Review | Acceptable | Unsatisfactory | Acceptable | Good | The authors combined pre-trained BERT models for text feature extraction, word, and label embeddings with a Self-Interaction Attention Mechanism |
| [50] 2022 | Multi30K | Good | Fair | Fair | Acceptable | The authors improved Multimodal Machine Translation models by training word embeddings on a monolingual corpus |

### 2.3.3 Transductive SVM-based Technique

By incorporating transduction principles, the technique broadens the capabilities of SVMs to handle partially labeled data. It prioritizes minimizing misclassifications for a set of test examples. The SVM tries to find the hyperplane that minimizes the number of misclassifications for the test examples, while maintaining separation between the classes.

*The rationale behind the usage of the technique:* To enhance the generalization accuracy of SVMs in text classification, a large margin hyperplane classifier is trained using labeled data. This enables the margin on test examples to become informative prior knowledge for learning, thereby improving performance. Moreover, integrating unlabeled data can further enhance the performance.

*Conditions for the optimal performance of the technique:* The use of One-Class SVM in transductive learning can enhance the technique. Furthermore, during training, an optimized concave-convex procedure can further improve the technique by iteratively solving a sequence of convex problems that approximate the concave function linearly around the solution of the previous convex problem.

*Limitations of the technique:* The performance of this technique is suboptimal when the dataset contains excessive noise, meaning that the target classes overlap. Moreover, due to its high training complexity, it is not ideal for classifying large datasets.

**Table 10:** Featuring and evaluating research papers that have employed Transductive SVM-based techniques.

| Paper/ Year | Dataset | Scalability | Interpretability | Accuracy | Efficiency | Description |
|---|---|---|---|---|---|---|
| [51] 2017 | Cifar10, MNIST | Acceptable | Fair | Good | Acceptable | The authors proposed a transductive SVM approach that uses the concave-convex mechanism for training |
| [52] 2007 | webspam-uk2006 | Unsatisfactory | Fair | Acceptable | Fair | The authors proposed a transductive SVM for multi-view spectral clustering, which finds the optimal clustering solution for all graphs |

### 2.3.4 Non-Transductive SVM-based Technique

The technique attempt to minimize the risk of misclassification on new, unseen data rather than just on the training set. They aim to not only separate the categories but also to maximize the margin between them in the feature space, leading to better generalization on unseen data.

*The rationale behind the usage of the technique:* Non-transductive SVMs are adept at managing sparse text data by prioritizing informative support vectors, making them suitable for text with many zero-valued features. These SVMs are optimized for fixed, specific test datasets, aligning with text mining's need for precise classification without generalizing, thus minimizing overfitting.

*Conditions for the optimal performance of the technique:* (1) Fine-tuning parameters like the regularization parameter (C) and kernel-specific parameters (e.g., gamma for RBF) through cross-validation can significantly enhance SVM performance, (2) Employing PCA or LDA to lower feature space dimensionality helps preserve critical features, improving efficiency and potentially performance, and (3) SVMs excel when data points are distinctly separable with a wide margin; this often requires diligent preprocessing and feature engineering.

*Limitations of the technique:* (1) Non-transductive SVMs cannot use unlabeled data, limiting performance in tasks where such data is plentiful, (2) Their performance greatly depends on hyperparameters, making optimal configuration resource-intensive, (3) They may perform poorly on imbalanced datasets, leading to models biased towards the majority class, and (4) Assumes data is linearly separable, which isn't always true, posing challenges in selecting and tuning the appropriate kernel.

**Table 11:** Featuring and evaluating research papers that have employed Non-Transductive SVM-based techniques.

| Paper/ Year | Dataset | Scalability | Interpretability | Accuracy | Efficiency | Description |
|---|---|---|---|---|---|---|
| [53] 2022 | Author Collected | Good | Acceptable | Good | Fair | The authors proposed a composite kernel function and non-transductive SVM approach for sentiment analysis of Indian language data. They combined multiple single kernel functions by taking a weighted summation |
| [54] 2018 | Author Collected | Unsatisfactory | Fair | Acceptable | Unsatisfactory | They proposed a non-transductive SVM method for feature extraction in English text data in NLP. They applied Gaussian Mixture for clustering the reduced data |

### 2.3.5 Applications of Research Fields Utilizing Feature-based, Transductive SVM, and Non-Transductive SVM Sub-Techniques

- **Anomaly Detection:** Feature-based techniques identify anomalies by extracting significant features from data, while Transductive SVM leverages both labeled and unlabeled data to improve anomaly detection in sparse datasets, and Non-Transductive SVM focuses on labeled data to classify anomalies effectively.
- **Rare Event Prediction:** Transductive SVM is particularly useful in predicting rare events by using limited labeled data and enhancing it with unlabeled data, while feature-based techniques identify critical predictors, and Non-Transductive SVM classifies these events based on the extracted features from labeled datasets.
- **Fraud Detection:** Feature-based methods extract key attributes indicative of fraudulent behavior, Transductive SVM enhances detection by utilizing both labeled and unlabeled transaction data, and Non-Transductive SVM classifies transactions as fraudulent or not based on labeled features.
- **Medical Diagnosis:** Feature-based techniques focus on extracting relevant medical indicators, Transductive SVM leverages both labeled and unlabeled patient data to improve diagnostic accuracy, and Non-Transductive SVM classifies patient conditions using labeled diagnostic features.
- **News Classification:** Feature-based approaches extract key terms and topics from news articles, Transductive SVM utilizes partially labeled datasets to improve classification accuracy, and Non-Transductive SVM classifies news articles based on fully labeled data and extracted features.
- **Sentiment Analysis:** Feature-based techniques identify sentiment-related features in text, Transductive SVM uses both labeled and unlabeled data to refine sentiment classification, and Non-Transductive SVM classifies sentiments using only labeled data.
- **Named Entity Recognition:** Feature-based methods extract entities by identifying distinctive features, Transductive SVM enhances recognition by using a combination of labeled and unlabeled text, and Non-Transductive SVM classifies entities based on labeled data and extracted features.
- **Text Classification:** Feature-based techniques extract important textual features, Transductive SVM leverages unlabeled data to improve classification models, and Non-Transductive SVM classifies text solely based on labeled features.
- **Information Retrieval:** Feature-based approaches enhance retrieval by identifying and weighting important features in queries and documents, Transductive SVM improves retrieval accuracy by using both labeled and unlabeled data, and Non-Transductive SVM retrieves information based on labeled data and extracted features

# 3 Artificial Intelligence (AI) Research Field-Based Category

## 3.1 Traditional Methods Sub-Category

### 3.1.1 Decision Tree Classification Technique

A decision tree is a visual representation that uses internal nodes to represent tests on attributes, branches to represent outcomes, and leaf nodes to represent class labels. The path from the root to the leaf provides classification rules. Fig. 2 demonstrates this procedure.

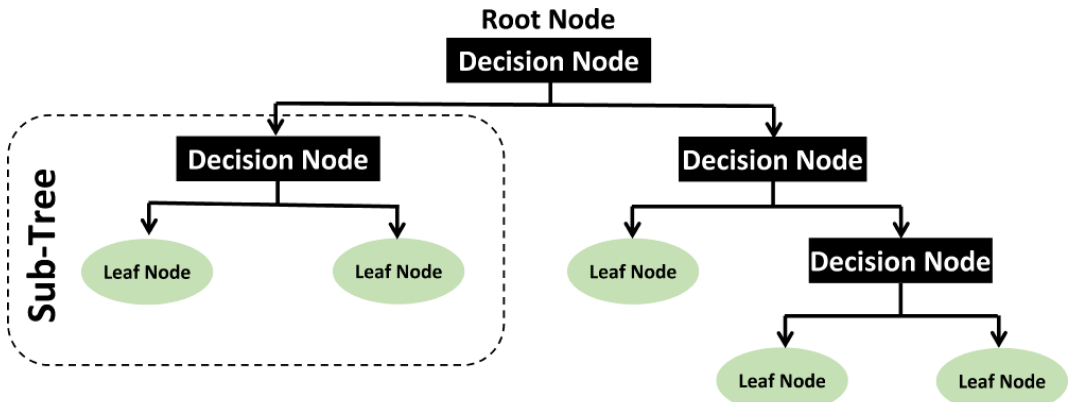

**Fig. 2:** The procedure of Decision Tree is illustrated in the figure.

#### *3.1.1.1 XGBoost-Based Sub-Technique*

To predict the target variable accurately, this technique employs supervised learning algorithms in the Gradient Boosting XGBoost framework to combine simpler models. It includes hyperparameters: tree depth (to specify the maximum depth of each decision tree), learning rate, and regularization parameters, which can be fine-tuned to improve the performance.

*The rationale behind the technique:* By utilizing sequentially built and parallel tree boosting, the technique enables the development of distributed gradient-boosted decision trees (GBDTs) with shallow depth. These GBDTs are both efficient and effective in accurately classifying text, as well as handling missing data within the text. Moreover, parameter optimization or tuning is not required in the technique.

*Conditions for the optimal performance of the technique:* Integrating weighted quantile sketch and sparsity can boost the accuracy of approximate tree learning in XGBoost's end-to-end tree boosting, thereby enhancing the technique. Moreover, combining XGBoost with the CHI statistical test can improve the technique by selecting the most relevant features. Also, integrating XGBoost with IDF can further enhance the technique's performance.

*Limitations of the technique:* The technique faces significant challenges when dealing with sparse and unstructured data, which can lead to suboptimal performance. Outliers can also pose a problem as each classifier must rectify the errors made by its preceding learners. Additionally, scalability is a concern as the overall approach is not easily scalable.

Table 12: Featuring and evaluating research papers that have employed XGBoost-based techniques.

| Paper/ Year | Dataset | Scalability | Interpretability | Accuracy | Efficiency | Description |
|---|---|---|---|---|---|---|
| [55] 2016 | Allstate-10K | Unsatisfactory | Good | Good | Unsatisfactory | They improved the accuracy of approximate tree learning in XGBoost by utilizing sparsity and weighted quantile sketch. They used a cache-aware block structure |
| [56] 2021 | Author Collected | Unsatisfactory | Acceptable | Acceptable | Unsatisfactory | The authors used XGBoost to classify scientific conference data in Indonesia |
| [57] 2022 | Author Collected | Unsatisfactory | Acceptable | Acceptable | Fair | They presented a technique for identifying technical debt in software development using XGBoost and the chi-square (CHI) statistical test |
| [58] 2021 | Author Collected | Fair | Acceptable | Good | Unsatisfactory | The authors used XGBoost to classify individuals' proactive personalities based on self-reported data |
| [59] 2020 | Nazario Phishing | Unsatisfactory | Fair | Good | Fair | The authors combined Mutual Information and CHI statistics with XGBoost to improve dimensionality reduction |

### 3.1.1.2 *LightGMB-Based Technique*

Using gradient boosting with LightGBM and a histogram-based decision tree learning approach, the technique discretizes continuous features into bins and selects the leaf node that minimizes the loss the most. The significance of gradient instances on information gain is relative to their size, with larger instances carrying a greater weight, and smaller instances having reduced importance.

*The rationale behind the usage of the technique:* Vertical growth of decision trees using a leaf-wise split approach has the potential to improve text classification accuracy. Incorporating Exclusive Feature Bundling and Gradient-Based One-Side Sampling techniques can increase algorithm efficiency and accuracy. Randomly dropping instances with small gradients while keeping instances with significant gradients helps to maintain accuracy.

*Conditions for the optimal performance of the technique:* The performance of a classification model is evaluated using the log loss function, and integrating a custom log loss function with LightGBM can enhance the accuracy of the technique. Customizing the log loss function enables better control over the learning process, allowing for optimal model performance to be achieved.

*Limitations of the technique:* It can suffer overfitting, which occurs when the tree becomes too deep and complex, making it difficult to interpret and apply to new data. This happens when a tree is split at the leaf level, as it leads to the generation of excessively complex trees. Such trees may capture the noise and idiosyncrasies in the training data, making them less effective at predicting new observations.

Table 13: Featuring and evaluating research papers that have employed LightGMB-based techniques.

| Paper/ Year | Dataset | Scalability | Interpretability | Accuracy | Efficiency | Description |
|---|---|---|---|---|---|---|
| [60] 2017 | Allstate, KDD, Flight Delay | Good | Good | Good | Good | The authors introduced LightGBM, an innovative Gradient Boosting Decision Tree (GBDT). This algorithm employs two crucial techniques: Exclusive Feature Bundling and Gradient-based One-Side Sampling |
| [61] 2021 | Author Collected | Fair | Fair | Acceptable | Acceptable | The authors utilized LightGBM to create a predictive model that forecasts trading risks and market volatility |
| [62] 2022 | EMBER, FFRI | Acceptable | Acceptable | Good | Good | The authors used a LightGBM-based method to examine malware classification accuracy, employing a custom log loss function to regulate the learning process |
| [63] 2021 | Exasens | Fair | Acceptable | Good | Fair | The authors used LightGBM in an ensemble method to detect diseases by assigning a significance score to each feature and sorting them accordingly |

### 3.1.1.3 Applications of Research Fields Utilizing XGBoost and LightGBM Sub-Techniques

- **Credit scoring**: XGBoost and LightGBM can build highly accurate models for credit scoring by efficiently handling large datasets and capturing complex patterns in financial data. These gradient boosting algorithms provide robust predictions by combining multiple weak learners to minimize errors.
- **Customer churn prediction**: XGBoost and LightGBM can predict customer churn by analyzing historical customer data and identifying patterns that lead to churn. Their ability to handle large-scale data and complex interactions makes them ideal for developing precise churn prediction models.
- **Sales forecasting**: XGBoost and LightGBM can forecast sales by analyzing past sales data and considering various factors such as seasonality, promotions, and economic conditions. Their fast training and high accuracy enable businesses to make reliable sales predictions.
- **Medical diagnosis**: XGBoost and LightGBM can assist in medical diagnosis by analyzing patient data, including medical history and lab results, to identify patterns associated with specific diseases. Their ability to handle heterogeneous data and complex relationships enhances diagnostic accuracy.
- **Financial risk assessment**: XGBoost and LightGBM can assess financial risk by modeling the relationships between various financial indicators and risk outcomes. Their strength in capturing non-linear relationships and interactions ensures accurate risk predictions.
- **News classification**: XGBoost and LightGBM can classify news articles by learning from labeled data and identifying relevant features for each category. Their efficiency and scalability allow for fast processing and accurate classification of large news datasets.
- **Named Entity Recognition (NER)**: While primarily used for structured data, XGBoost and LightGBM can be applied

to NER by converting text data into numerical features, though they may not be as effective as deep learning models designed for this task. They can still provide decent performance when combined with feature engineering techniques.

- **Text classification**: XGBoost and LightGBM can classify text by transforming textual data into numerical features using techniques like TF-IDF or word embeddings. These algorithms excel in handling high-dimensional data and capturing feature interactions, resulting in accurate text classification models.

### 3.1.2 Probabilistic Graphical Model Classification Technique

Probabilistic Graphical Model-Based text classification is a method that utilizes probabilistic graphical models to model the dependencies between the words in a text document. PGM-Based classification methods involve constructing a graphical model from a training set of labeled documents, where the nodes in the graph represent words or phrases, and the edges represent the dependencies between them. Fig. 3 depicts this process.

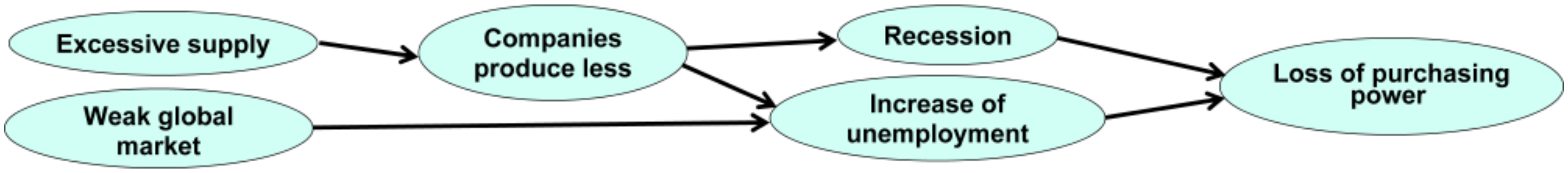

**Fig. 3**: The figure illustrates an example of the probabilistic graphical model process.

#### *3.1.2.1 Naïve Bayes Classifier Sub-Technique*

This technique utilizes Naïve Bayes, a probabilistic classifier, to classify instances represented as feature vectors. The classification process involves assigning class labels from a predetermined set to the problem instances.

*The rationale behind the usage of the technique:* The training process is expedited by solely computing the probability of each class and the probability of each class given distinct input values. Moreover, the independence of all features results in Naive Bayes being exceptionally swift.

*Conditions for the optimal performance:* Assigning feature-specific weights to each class can improve the technique, whereby the conditional probabilities of the text classifier are calculated by utilizing feature-weighted frequencies obtained from the training data.

*Limitations of the technique:* It would assign zero probability to any categorical variable in the test data set that is absent in the training data, rendering it impossible to make any predictions. The naive form of Naive Bayes lacks support for parallel computation.

**Table 14:** Featuring and evaluating research papers that have employed Naïve Bayes-Based Classifier

| Paper/ Year | Dataset | Scalability | Interpretability | Accuracy | Efficiency | Description |
|---|---|---|---|---|---|---|
| [64] 2022 | UCI, Wholesale Customers | Good | Good | Good | Acceptable | The authors developed a new attribute weighting technique to enhance Naive Bayesian Classification performance. This technique uses non-linear optimization to assign attribute weights |
| [65] 2020 | Author Collected | Acceptable | Acceptable | Good | Fair | The authors proposed an under-sampling technique using active selection to create a modified training set for Naive Bayes classifiers |
| [66] 2020 | Fbis, trl, ohscal | Unsatisfactory | Fair | Acceptable | Good | They presented a class-specific feature weighting approach for Multinomial Naive Bayes classifiers that assigns weights to each feature for each class |
| [67] 2016 | REUTERS | Unsatisfactory | Fair | Fair | Acceptable | The authors implemented a class-specific feature subset selection approach for text categorization using a Bayesian classification rule |

#### *3.1.2.2 Naïve Bayes Transfer Classifier Sub-Technique*

In the technique, the EM-based Naive Bayes classifiers estimate the initial probabilities from a labeled dataset and then use an EM algorithm to revise them to account for a different distribution of unlabeled test data.

*The rationale behind the usage of the technique:* Applying the EM algorithm to derive an initial model using labeled data from the Du distribution leads to the algorithm iteratively converging towards a locally optimal model that aligns with the same Du distribution.

*Conditions for the optimal performance of the technique:* The performance of text classification can be enhanced by utilizing both labeled and unlabeled data to optimize semi-supervised label propagation through a combination of the Naïve Bayes Transfer Classifier and Cross-Domain Kernel Induction.

*Limitations of the technique:* The technique assumes that features are not interdependent, which prevents it from learning the relationship between features. Consequently, it can be categorized as a low variance high bias classifier due to its conditional independence.

**Table 15:** Featuring and evaluating research papers that have employed Naïve Bayes Transfer Classifier-Based techniques.

| Paper/ Year | Dataset | Scalability | Interpretability | Accuracy | Efficiency | Description |
|---|---|---|---|---|---|---|
| [68] 2018 | heart-statlog, segment, vowel | Acceptable | Acceptable | Good | Unsatisfactory | They proposed TrResampling, an ensemble transfer learning method that incorporates bagging, MultiBoosting, and weighted resampling with TrAdaBoost. The learners of TrResampling are NBTC and decision tree. |
| [69] 2007 | SRAA2, Reuters | Fair | Unsatisfactory | Acceptable | Acceptable | They proposed the Naive Bayes Transfer Classifier for text classification. NBTC utilizes an Expectation-Maximization algorithm Naive Bayes classifier to estimate the initial probabilities based on the labeled dataset's distribution. |
| [70] 2018 | Movielens, Netflix | Fair | Fair | Acceptable | Acceptable | They combined NBTC with Cross-Domain Kernel Induction to optimize the semi-supervised label propagation to enhance text classification. |

#### 3.1.2.3 Applications of Research Fields Utilizing Naïve Bayes and Naïve Bayes Transfer Sub-Techniques

- **Spam filtering**: Naïve Bayes can classify emails as spam or not spam by calculating the probability of an email being spam based on the frequency of certain words and features. Naïve Bayes Transfer enhances this by leveraging pre-trained models on related tasks to improve accuracy and adaptability to new spam trends.
- **Topic categorization**: Naïve Bayes can categorize documents into topics by using the probability of words belonging to specific topics, assuming word independence. Naïve Bayes Transfer can utilize knowledge from similar categorization tasks to improve performance on new but related topics.
- **Real-time email classification**: Naïve Bayes can classify emails in real-time by quickly computing the probability of an email belonging to various categories based on its content. Naïve Bayes Transfer allows the system to adapt to new email patterns by incorporating pre-learned information from similar classification tasks.
- **Adaptive filtering systems**: Naïve Bayes can be used in adaptive filtering systems to continuously learn and update classification models based on new data, maintaining high accuracy. Naïve Bayes Transfer enhances this adaptability by transferring knowledge from previously trained models to quickly adjust to changing data patterns.
- **News classification**: Naïve Bayes can classify news articles by calculating the likelihood of an article belonging to a certain category based on word frequencies. Naïve Bayes Transfer can improve classification accuracy by leveraging pre-trained models on related news datasets.
- **Named Entity Recognition (NER)**: While Naïve Bayes is less commonly used for NER, it can still identify named entities by probabilistically determining the likelihood of words being entities based on labeled training data. Naïve Bayes Transfer can enhance this by utilizing knowledge from pre-trained NER models on similar corpora.
- **Text classification**: Naïve Bayes can perform text classification by calculating the probability of a text belonging to different categories based on word occurrences. Naïve Bayes Transfer boosts this capability by incorporating insights from pre-trained models on related classification tasks, improving accuracy and robustness.

### 3.1.3 Nearest Neighbor Classification Technique

The technique determines a sample's classification by examining the K most similar samples in the feature space. If a sample's K nearest neighbors is predominantly from a particular category, it is assigned to that category. Some techniques compute similarities between a given document's feature vector and each document vector in the training set to identify the K documents with the highest similarity to the new document. Then, the new document's class is determined based on the classes of these K documents. Fig. 4 demonstrates this.

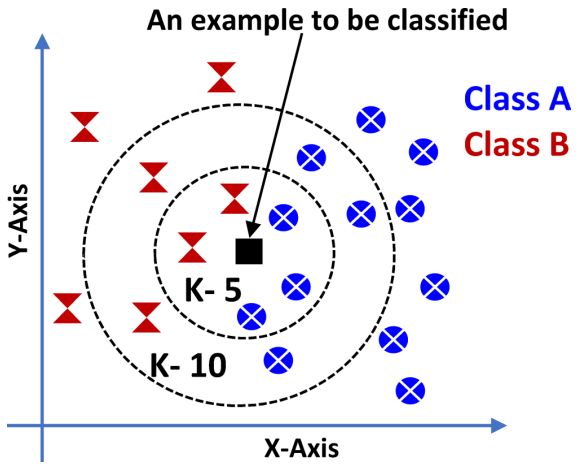

**Fig. 4:** The figure depicts the process of the K-Nearest Neighbors-based classifier.

#### *3.1.3.1 K-Nearest Neighbors Sub-Technique*

The technique relies on proximity to classify data by taking *k* nearest instances from a dataset as input and generating a class membership as output. This is achieved by assigning each point to the most common class among its *k* nearest neighbors.

*The rationale behind the usage of the technique:* The fundamental idea is that objects in close proximity within a dataset tend to exhibit similarities. Consequently, utilizing a classification algorithm that considers distance and normalizing the training data can lead to a considerable improvement in the algorithm's accuracy.

*Conditions for the optimal performance of the technique:* Utilizing evolutionary algorithms to optimize feature scaling can help overcome the challenge posed by noisy or irrelevant features that can negatively impact the accuracy of the k-NN algorithm. Also, scaling the features according to their mutual information with the training classes can improve the accuracy.

*Limitations of the technique:* The effectiveness of the technique can be severely impacted by the presence of inconsistent feature scales and noisy or irrelevant features. Additionally, determining the appropriate value of k can be a challenging task, and the technique is computationally inefficient.

**Table 16:** Featuring and evaluating research papers that have employed K-Nearest Neighbor-based techniques.

| Paper/ Year | Dataset | Scalability | Interpretability | Accuracy | Efficiency | Description |
|---|---|---|---|---|---|---|
| [71] 2020 | Moons, Blobs | Unsatisfactory | Good | Acceptable | Unsatisfactory | The authors suggested a spectral clustering method that utilized k-nearest neighbors to construct a graph of nearest neighbors |
| [72] 2020 | Synthetic datasets | Unsatisfactory | Acceptable | Good | Acceptable | The authors proposed a k-nearest neighbor (kNN) approach for managing large datasets in cyber-physical-social systems |
| [73] 2020 | Synthetic datasets | Unsatisfactory | Acceptable | Acceptable | Fair | The authors proposed a density peaks clustering method that divides kNN into multiple layers to handle varying density datasets |
| [74] 2023 | MNIST, USPS | Unsatisfactory | Fair | Acceptable | Fair | The authors combined kNN and CNN through learning a metric that considers the relationships between neighboring instances and features |
| [75] 2020 | MNIST, USPS | Unsatisfactory | Acceptable | Acceptable | Acceptable | They investigated the effectiveness of using kNN search when the feature space is represented by the activations of the final layers of a neural network |

### 3.1.3.2 *Weighted K-Nearest Neighbors Sub-Technique*

The technique factors in feature uncertainties by assessing the similarity between data points using distance measurements. The selection of the distance metric is determined through learning, which creates a metric tailored to the specific task.

*The rationale behind the usage of the technique:* A model can be enhanced by assigning weights to the nearest $k$ points, with greater weight given to closer points and less weight given to those farther away. A kernel function can be applied to the weighted KNN classifier, which decreases in value as distance increases.

*Conditions for the optimal performance of the technique:* The technique can be improved by computing the density coefficients of data points for each class in the training set individually. This is because noisy data points, which are farther away from the set center than other data points, have lower density.

*Limitations of the technique:* The distance measurement can result in suboptimal performance when dealing with complex datasets that have multiple scales, varying densities, or high dimensionality. Also, determining the cutoff distance is typically challenging since the range of each attribute is often unknown.

**Table 17:** Featuring and evaluating research papers that have employed Weighted K-Nearest Neighbor-based techniques.

| Paper/ Year | Dataset | Scalability | Interpretability | Accuracy | Efficiency | Description |
|---|---|---|---|---|---|---|
| [76] 2018 | Author Collected | Fair | Good | Good | Fair | The authors proposed a weighted KNN approach for local learning in Stack Overflow data. The approach involved clustering the data and constructing separate local models for each cluster |
| [77] 2014 | WSDL | Unsatisfactory | Acceptable | Fair | Unsatisfactory | The authors presented a service retrieval technique that utilized a weighted KNN to address the challenge of feature extraction |
| [78] 2016 | Author Collected | Unsatisfactory | Acceptable | Acceptable | Fair | They proposed a method that employed a weighted KNN approach, which involved computing K cluster centers of the minority class using K-means and generating multiple sample points for each center |

### 3.1.3.3 Applications of Research Fields Utilizing KNN and Weighted KNN Sub-Techniques

- **Recommender systems**: KNN can provide recommendations by finding and suggesting items similar to a user's past preferences based on nearest neighbors. Weighted KNN enhances this by giving more influence to closer neighbors, improving the relevance of the recommendations.
- **Personalized medicine**: KNN can predict patient outcomes or suggest treatments by comparing a patient's characteristics with those of similar patients. Weighted KNN can refine these predictions by weighting the influence of closer, more similar patients more heavily.
- **Anomaly detection**: KNN detects anomalies by identifying data points that have few or no similar neighbors, indicating they are outliers. Weighted KNN improves the sensitivity of anomaly detection by considering the distance and assigning higher weights to closer neighbors.
- **Geographic profiling of crimes**: KNN can analyze crime locations to predict future crime hotspots by identifying patterns based on the nearest crime locations. Weighted KNN enhances the accuracy by giving more importance to closer crime locations, providing more precise geographic profiles.
- **Location-based services**: KNN can provide services such as navigation or recommendations by finding the nearest points of interest based on a user's location. Weighted KNN improves these services by prioritizing closer locations, ensuring more relevant results.
- **News classification**: KNN can classify news articles by finding and comparing them with similar, already categorized articles based on content similarity. Weighted KNN improves classification accuracy by giving more weight to closer, more similar articles.
- **Named Entity Recognition (NER)**: While less common for NER, KNN can still identify entities by comparing text with labeled examples and identifying similar patterns. Weighted KNN enhances entity recognition by giving more importance to closer matches, improving accuracy.
- **Text classification**: KNN classifies text by finding the closest examples in a labeled dataset and assigning the most common category among the neighbors. Weighted KNN improves classification by assigning higher weights to closer neighbors, resulting in more accurate text categorization.

## 3.2 Deep Learning-Based Classification Sub-Category

Text classification is a common application of deep learning, which involves using artificial neural networks to learn high-level features from input data. This approach involves intricate transformations, similar to the human brain, to train the model effectively. To do so, appropriate input word vectors are prepared and given to the neural networks, considering various factors such as data type, supervision, and balance. After training, the model's efficiency is measured based on downstream tasks like sentiment classification, question answering, and event prediction. Fig. 3 demonstrates this process. Deep learning models have outperformed traditional models by detecting complex patterns, resulting in precise and efficient models. It has become a valuable tool for researchers looking to create accurate and efficient models for text classification. Fig. 5 demonstrates the procedure of deep learning.

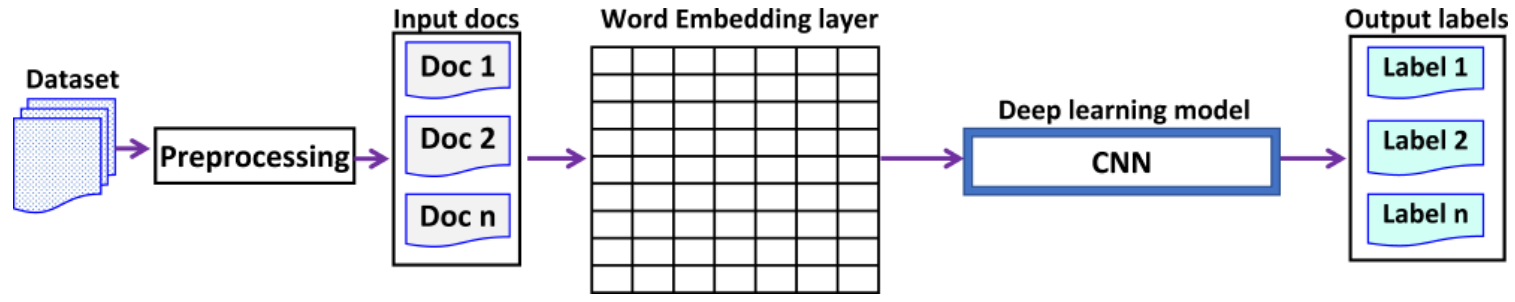

**Fig. 5:** The figure demonstrates the process of deep learning.

### 3.2.1 Residual Network-based Technique

This technique leverages residual neural networks (RNNs) to improve text classification by extracting advanced text representations and addressing vanishing gradients, a notable challenge in deep neural network training. By using a deep residual network, it supports incremental text classification and the integration of new data without full retraining. RNNs are adept at identifying the subtle nuances in text, revealing the semantic structure of language for more accurate analysis and interpretation of complex datasets. Enhancements like Rectified Linear (ReLU) nonlinearities and batch normalization between layers, especially in double- or triple-layer skips, accelerate training convergence and boost the network's robustness, enabling precise handling of classifications.

*The rationale behind the usage of the technique:* By mitigating accuracy saturation through the resolution of vanishing gradients, neural networks can be optimized and trained more quickly and effectively. This helps to improve their performance on a wide range of tasks, from image and speech recognition to NLP.

*Conditions for the optimal performance of the technique:* When a single nonlinear layer is skipped, this can allow the network to capture more complex relationships between the input features. By allowing information to bypass a nonlinear layer, the network can effectively create a shortcut that allows it to access information from a previous layer that may have been lost.

*Limitations of the technique:* (1) comprise of high computational complexity, (2) susceptibility to overfitting (when the model is too complex or when there is insufficient training data), (3) high memory usage, (4) difficulty during training, and limited scalability.

**Table 18:** Featuring and evaluating research papers that have employed Residual Network-based techniques.

| Paper/ Year | Dataset | Scalability | Interpretability | Accuracy | Efficiency | Description |
|---|---|---|---|---|---|---|
| [79] 2016 | CIFAR-10, CIFAR-100 | Good | Unsatisfactory | Good | Fair | They introduced a deep learning method for text and image classification that employs deep residual networks for efficient information propagation. |
| [80] 2020 | SST-1, SST-2, TREC | Good | Fair | Acceptable | Acceptable | The authors presented a stacked residual network method for text classification that uses a cross-layer attention mechanism to filter linguistic features and supervise lower-level features. |
| [81] 2017 | Ren, Zhou datasets | Acceptable | Fair | Good | Fair | They combined residual networks with a TSCD layer to extract Chinese text, utilizing the residual mechanism to improve text classification. |

### 3.2.2 Deep Pyramid-based Technique

This technique starts by inputting a sentence into the text region embedding layer, where word embedding is utilized to generate vector representations for each word. Following this, two convolutional layers are applied to further process the data. These algorithms leverage a pyramid network structure along with the regional embedding method to capture contextual semantics and long-range dependencies among words effectively. The deep pyramid CNN architecture is designed to represent global information and long-range connections efficiently. Concurrently, the regional embedding technique models the relationships between words in their local surroundings. By integrating these methods, the algorithms can capture the rich semantic information of language.

*The rationale behind the usage of the technique:* The "pyramid" aids in identifying long-range relationships within the text, while downsampling captures more information. Additionally, it has low computational complexity, fewer parameters, and necessitates less memory space on disk.

*Conditions for the optimal performance of the technique:* The incorporation of a residual connection into the pyramid CNN addresses weight loss and gradient dispersion problems that occur with increased network depth, leading to improved accuracy. Residual connection allows network to learn and identity mapping.

*Limitations of the technique:* Deeper network depth can cause weight loss and gradient dispersion issues in this technique, and it also requires a substantial amount of training data.

**Table 19:** Featuring and evaluating research papers that have employed Deep Pyramid Convolutional Neural Network-based techniques.

| Paper/ Year | Dataset | Scalability | Interpretability | Accuracy | Efficiency | Description |
|---|---|---|---|---|---|---|
| [82] 2022 | Author Collected | Good | Unsatisfactory | Acceptable | Good | They introduced Deep Pyramid Convolutional Neural Network that uses convolutional operators to learn features at various levels and captures global text features by increasing the network depth, while the pyramid structure discovers long-range relationships. |
| [83] 2022 | Hotel review data set | Acceptable | Unsatisfactory | Acceptable | Acceptable | They captured local spatial features of sentences and analyzed contextual semantic association information, using sentences as random vectors. |

### 3.2.3 TextCNN-based Technique

In a 4-layer architecture, this technique utilizes a one-dimensional convolutional layer and a sequential maximum pooling layer along with an embedding layer.

*The rationale behind the usage of the technique:* Modifying the height of the convolution kernel enables a model to process different temporal information from the vocabulary, leading to improved text comprehension. Additionally, the embedding layer can have fewer parameters without significant accuracy loss.

*Conditions for the optimal performance of the technique:* Incorporating chunk-based max-pooling into the model can further enhance the pooling layer and preserve more text features obtained through convolution, leading to improved performance.

*Limitations of the technique:* A potential limitation of the technique is the presence of irrelevant words among semantically related words. Moreover, a significant amount of sample data is required for effective training of the model.

**Table 20:** Featuring and evaluating research papers that have employed TextCNN-based techniques.

| Paper/ Year | Dataset | Scalability | Interpretability | Accuracy | Efficiency | Description |
|---|---|---|---|---|---|---|
| [84] 2014 | SST-1 &2, MR, TREC | Fair | Acceptable | Good | Acceptable | They presented a TextCNN approach that employs a convolutional neural network based on the word2vec to classify sentences. It used fixed word vectors initially, but better outcomes were obtained by fine-tuning them. |
| [85] 2021 | Movie Review dataset | Unsatisfactory | Acceptable | Acceptable | Fair | They combined TextCNN with SVM for text sentiment analysis, where TextCNN extracted features fed into the SVM classifier, replacing the SoftMax layer. The classifier uses TextCNN's penultimate feature vector. |
| [86] 2020 | Ransomware samples | Fair | Fair | Good | Fair | They advanced the TextCNN method by integrating dynamic ransomware detector and enhancing the pooling layer using chunk max-pooling |

### 3.2.4 Applications of Research Fields Utilizing Residual Network Techniques, Deep Pyramid, and TextCNN Sub-Techniques

- **Autonomous driving**: Residual Networks (ResNet) can improve object detection and image segmentation in autonomous driving by allowing deeper networks to learn complex features without vanishing gradients. Deep Pyramid networks enhance this by combining multi-scale features for better detection of objects at various distances. TextCNN isn't typically used in autonomous driving but can help analyze textual data related to driving scenarios, like traffic reports.
- **Medical imaging**: ResNet can improve the accuracy of diagnosing diseases from medical images by learning complex patterns in the data. Deep Pyramid networks enhance medical image analysis by leveraging multi-scale features to detect abnormalities of various sizes. TextCNN can be applied to classify radiology reports and extract relevant medical information from text.
- **Hierarchical text classification**: ResNet can be adapted for hierarchical text classification by stacking layers that progressively capture higher-level features. Deep Pyramid networks can effectively handle varying text lengths and complexities by integrating multi-scale textual features. TextCNN can classify text hierarchically by capturing local features through convolutional layers, followed by pooling layers to aggregate information.
- **Multi-level sentiment analysis**: ResNet can analyze sentiments at different levels of text by learning hierarchical features from sentences to documents. Deep Pyramid networks can enhance sentiment analysis by considering sentiment cues at multiple scales within the text. TextCNN excels at capturing local sentiment features using convolutional filters, enabling multi-level sentiment classification.
- **News classification**: ResNet can classify news articles by learning hierarchical features from words to paragraphs, capturing complex patterns in news data. Deep Pyramid networks enhance news classification by integrating features from different text levels for a comprehensive understanding. TextCNN can classify news by extracting key features from text through convolutional layers, providing accurate categorization.
- **Named Entity Recognition (NER)**: ResNet can be adapted to identify entities by learning rich, hierarchical representations of text. Deep Pyramid networks enhance NER by leveraging multi-scale text features, improving entity recognition across different contexts. TextCNN can identify named entities by capturing local context through convolutional layers, providing effective recognition of entities within text.
- **Text classification**: ResNet can classify text by learning deep, hierarchical features, enabling accurate categorization of complex text data. Deep Pyramid networks enhance text classification by integrating multi-level text features for a more nuanced understanding. TextCNN effectively classifies text by capturing local patterns and features using convolutional filters, followed by pooling layers to consolidate information.

## 4. Experimental Evaluation

We experimentally evaluated and ranked the techniques discussed in this paper. To represent each technique, we selected an algorithm and executed it on a machine running Windows 10 Pro with an Intel(R) Core(TM) i7-6820HQ processor, 2.70 GHz CPU speed, and 16 GB RAM.

### 4.1 Methodology for Selecting a Representative Paper for each Sub-Technique and Ranking the Various Sub-Techniques under the Same Technique and Overall Research Field Sub-Category

After reviewing the papers that reported algorithms utilizing a particular sub-technique, we identified the most influential one. These selected algorithms serve as the representative sub-techniques. To determine the most significant paper among all papers reporting algorithms that use the same sub-technique, we assessed different factors, including how cutting-edge it is and its publication date. Table 2x shows the chosen paper for each technique.

We searched for the publicly available codes for the algorithms reported in the papers that we selected as representative of the techniques described in this survey. We could obtain codes for the following paper: [42[1]], [24[2]], [82[3]], [34[4]], [50[5]], [48[6]], [19[7]], [59[8]], [120[9]], [117[10]]. For the remaining representative papers, we developed our own implementations using TensorFlow [87] and trained them with the Adam optimizer [87]. TensorFlow's APIs enable users to create custom ML algorithms [88]. We utilized Python 3.6 as our development language and TensorFlow 2.10.0 as the backend for the models.

Specifically, we compared and ranked the methodology sub-techniques within the same methodology technique and within the same overall research field sub-category. We conduct experimental evaluations under the following five scenarios:

1. **Scenario 1: Comparing Sub-Techniques within Recurrent Neural Network sub-category of Data Science Category**: These sub-techniques are used for sentiment and sequence classification tasks, which are crucial for analyzing user-generated content. We applied them to the Yelp reviews dataset [9].
2. **Scenario 2: Comparing Sub-Techniques within Embedding sub-category of Data Science Category**: These sub-techniques are essential for handling text with varied formats and errors, which is crucial for news categorization. We utilized the BBC News dataset [7] for this purpose.
3. **Scenario 3: Comparing Sub-Techniques within Pre-Trained sub-category of Data Science Category**: These sub-techniques are used to measure performance in transfer learning for sentiment analysis, utilizing the SST-2 dataset [8].
4. **Scenario 4: Comparing Sub-Techniques within Traditional Methods of Artificial Intelligence Category**: We evaluate these sub-techniques for their ongoing relevance and limitations in contemporary text classification tasks, using the Reuters-21578 dataset [10].
5. **Scenario 5: Comparing Sub-Techniques within Deep Learning Methods of Artificial Intelligence Category**: These sub-techniques handle large-scale, diverse text data for sentiment analysis and product review classification, supporting commercial applications. We utilized the Amazon product reviews dataset [11] for this purpose.

### 4.2 Methodology for Conducting Evaluations and Experimental Setup

Table 21 provides a description of the five scenarios, including the selected papers representing the sub-techniques compared in each scenario, the dataset used for evaluation, the evaluation metrics, and the experimental parameters adopted for each scenario.

**Table 21:** Description of the five scenarios, including the selected papers representing the sub-techniques compared in each scenario, the dataset used for evaluation, the evaluation metrics, and the experimental parameters adopted for each scenario

| Scenario | Description | Compared Sub-Techniques | Dataset | Metrics | Experimental Parameters |
|---|---|---|---|---|---|
| **1** | Comparing **Recurrent Neural Network-Based Data Science sub-techniques**. | LSTM [18], Language Inference [23] | Yelp reviews dataset [9] | Accuracy, Precision, Recall, F1-Score | LSTM Units= 128; Dropout Rate = 0.5; |
| **2** | Comparing **Embedding-Based Data Science sub-techniques**. | Capsule [26], Bidirectional LSTM [30], Out-Of-Vocabulary [34], CNN [36] | BBC News dataset [7] | Accuracy, Precision, Recall, F1-Score | LSTM Units= 128; Dropout Rate = 0.5; Capsules = 10; Routing Iterations = 3 |
| **3** | Comparing **Pre-Trained-Based Data Science sub-techniques.** | BERT [39], ALBERT [43], Feature-based [46], Transductive SVM [51], Non-Transd. SVM [53] | SST-2 [8] | Accuracy, F1-Score | Model Version = Base version for comparability; Fine-tuning Epochs = 4 |
| **4** | Comparing **Traditional Artificial Intelligence sub-techniques**. | XGBoost [55], LightGMB [60], Naïve Bayes [64], Naïve Bayes Transfer [68], KNN [71], Weighted KNN [76] | Reuters-21578 dataset [10] | Accuracy, Precision, Recall, F1-Score | Max Depth = 6; Number of Estimators = 100; Alpha (Smoothing parameter) = 1.0 |
| **5** | Comparing **Deep Learning Artificial Intelligence sub-techniques**. | Residual Network [79], Deep Pyramid CNN [82], TextCNN [84] | Amazon product reviews [11] | Accuracy, Precision, Recall, F1-Score | Filter Sizes = 4; Number of Filters = 100 each size; Dropout Rate = 0.5 |

### 4.3 Scenario 1: Comparing Sub-Techniques within Recurrent Neural Network sub-category of Data Science Category

In this scenario, we evaluated and compared recurrent neural network-based data science sub-techniques for sentiment and sequence classification tasks, which are vital for analyzing user-generated content. These methods were assessed using the Yelp reviews dataset [9]. The experimental results are presented in Figure 6.

---

[1] https://github.com/dmlc/xgboost
[2] https://github.com/Microsoft/LightGBM
[3] https://github.com/tangjianpku/LINE
[4] https://github.com/google-research/bert
[5] http://mlcv.ogu.edu.tr/softwarertsvm.html
[6] https://github.com/akashgit/autoencoding_vi_for_topic_model
[7] https://sfb876.tu-dortmund.de/spectacl/index.html
[8] http://clair.si.umich.edu/corpora/citation_sentiment_umich.tar.gz
[9] https://www.isical.ac.in/%e2%88%bcbibl/results/redmica/redmica.html
[10] https://github.com/luoxudong/DDNM

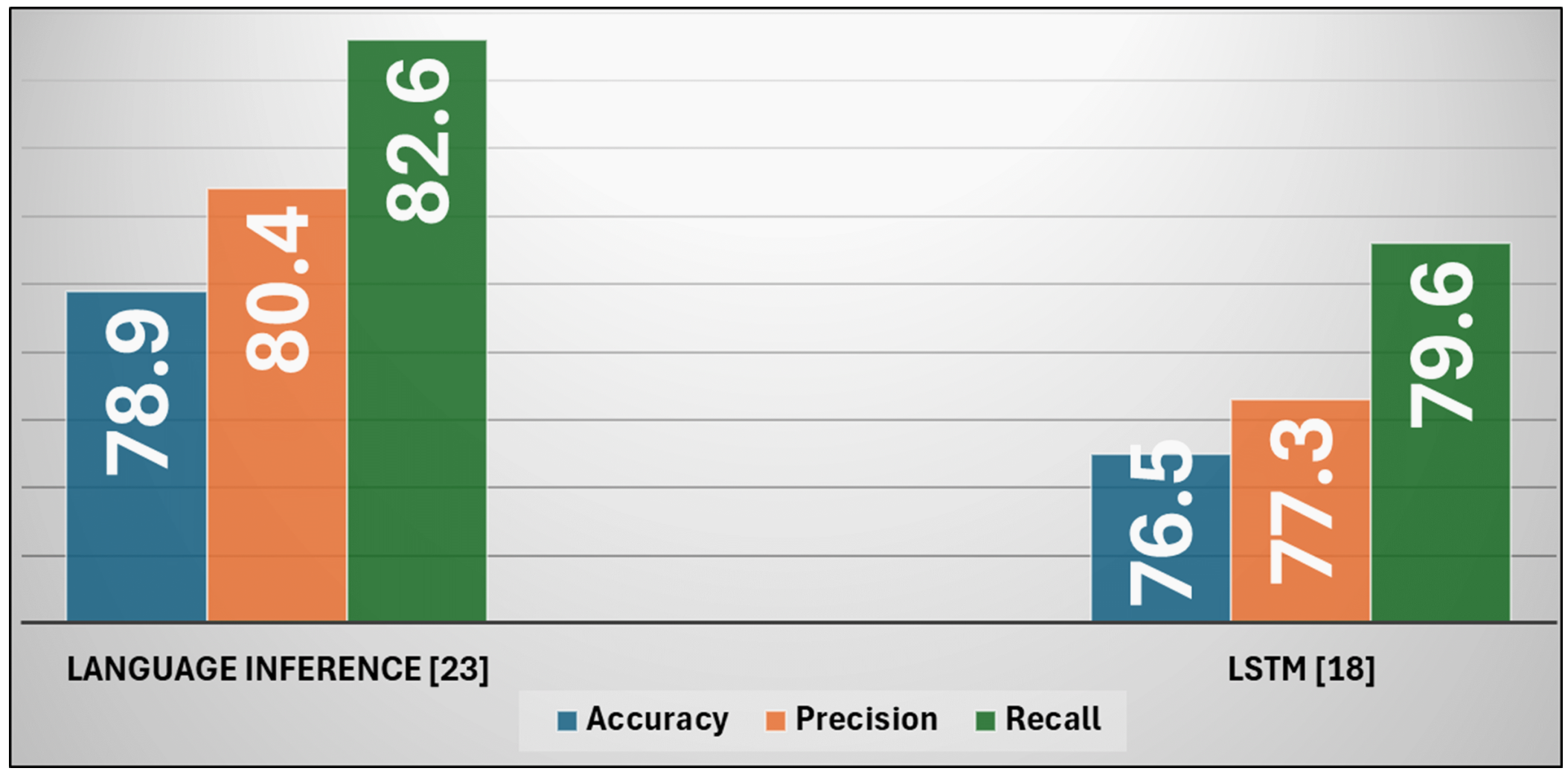

**Fig. 6:** Depicting the results of comparing Recurrent Neural Network-Based Data Science sub-techniques

Although these algorithms showed some ability to acquire grammatical connections by identifying relevant details at a later stage in a sequence and reconstructing disrupted input into coherent text, they could not accurately comprehend complex sentence structures or non-sequential input data. As a result, their accuracy levels remained moderate, as suggested by the experimental findings.

Language inference-based method, focused on understanding and predicting relationships within text. While traditional methods may struggle with context retention, advanced inference technique provided superior accuracy and understanding. Despite the computational intensity, the enhanced performance in capturing intricate language patterns makes LSTMs and advanced inference technique invaluable in natural language processing applications.

The LSTM model excelled in capturing long-range dependencies and context within sequences, resulting in acceptable accuracy. It was particularly effective in handling complex sentence structures and nuanced meanings, outperformed traditional methods. However, LSTMs demanded significant computational resources and longer training times, with performance heavily dependent on the quality and quantity of training data.

## 4.4 Scenario 2: Comparing Sub-Techniques within Embedding sub-category of Data Science Category

In this scenario, we evaluated and compared embedding-based data science sub-techniques designed to handle text with diverse formats and errors, which are essential for effective news categorization. The assessment was conducted using the BBC News dataset [7]. Figure 7 presents the experimental results.

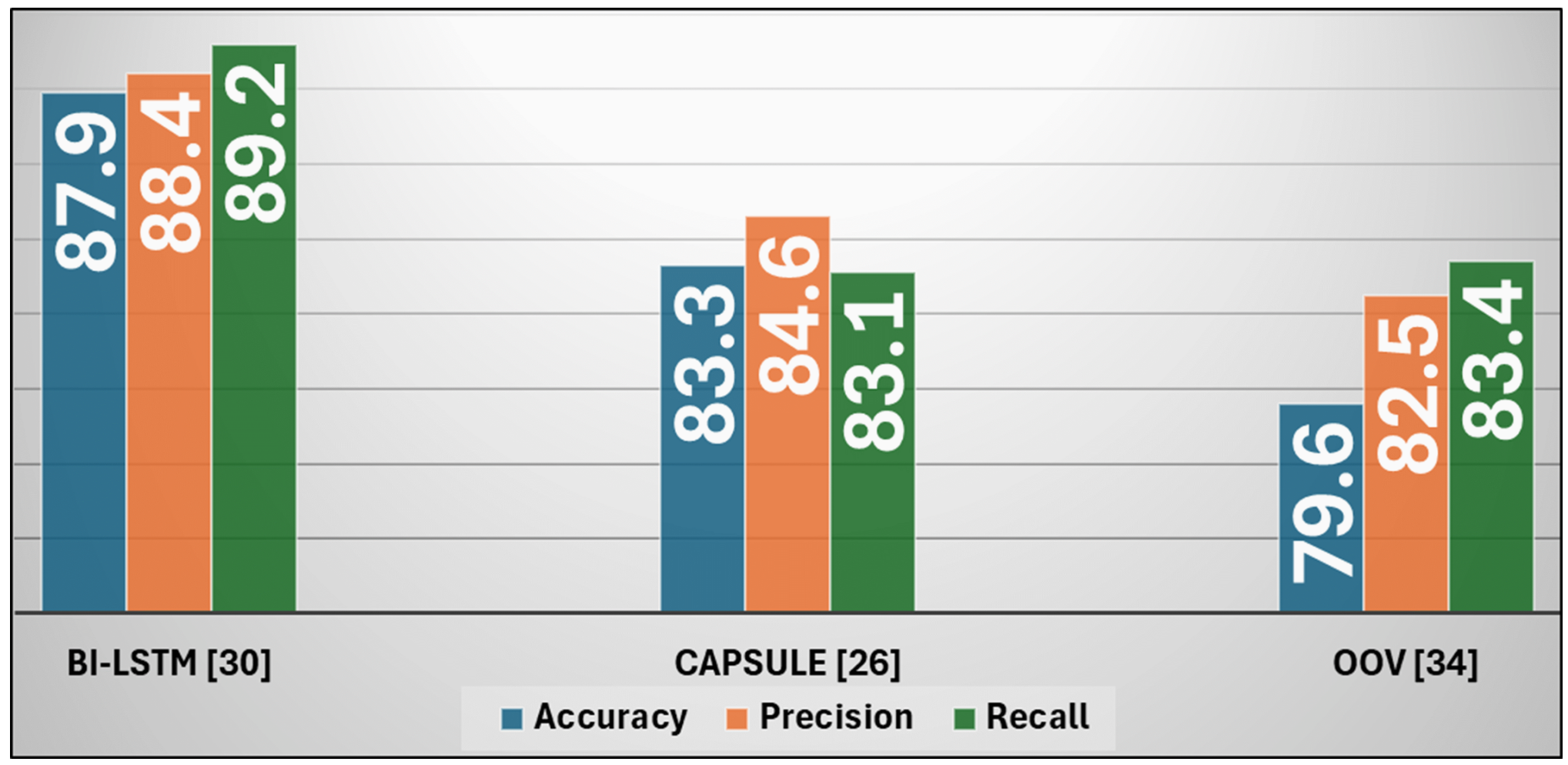

**Fig. 7:** Depicting the results of comparing Embedding-Based Data Science sub-techniques

Bi-LSTM demonstrated strong performance in capturing contextual information from text, effectively managing sequential data, and handling diverse text formats. This resulted in high accuracy and robustness in categorizing news articles. Conversely, Capsule Networks exhibited superior capabilities in recognizing hierarchical relationships within the text, which enhanced their ability to manage complex patterns and errors. Capsules showed resilience against variations and distortions in the text data.

Overall, while Bi-LSTM provided excellent context management and sequence learning, Capsule Networks offered advanced hierarchical pattern recognition, making them particularly effective in scenarios involving diverse and error-prone text. The OOV technique showed relatively modest results.

## 4.5 Scenario 3: Comparing Sub-Techniques within Pre-Trained sub-category of Data Science Category

In this scenario, we evaluated and compared pre-trained data science sub-techniques to measure performance in transfer learning for sentiment analysis. The assessment was conducted using the SST-2 dataset [8]. Figure 8 illustrates the experimental results.

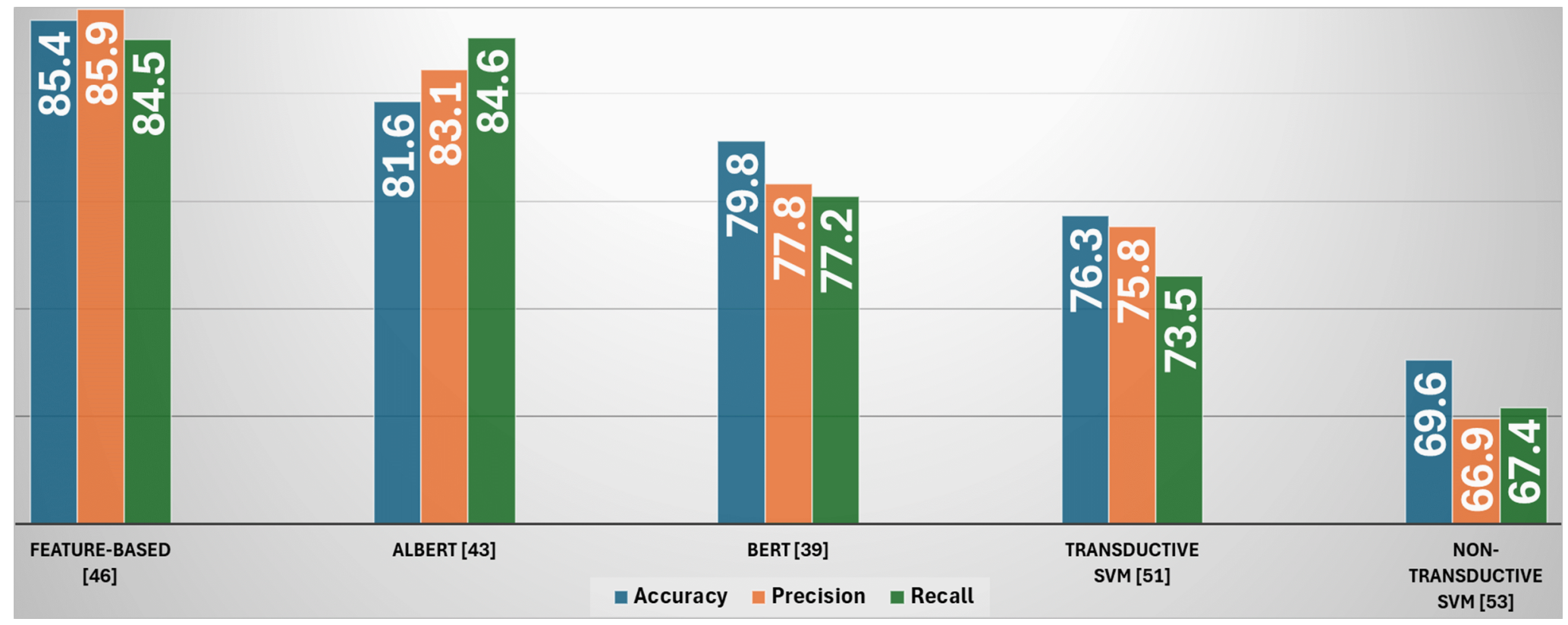

**Fig. 8**: Depicting the results of comparing Pre-Trained-Based Data Science sub-techniques

The feature-based method excelled by extracting meaningful features like term frequency, TF-IDF, and word embeddings, which capture important information from the text. It reduced dimensionality, allowing the method to focus on the most relevant aspects, thus improving efficiency and accuracy. It effectively handled sparse data and could combine multiple feature types to provide a comprehensive text representation.

Despite capturing word contexts and analyzing significant contextual information, BERT and ALBERT achieved good accuracy scores. BERT excelled due to its bidirectional context understanding but required substantial computational resources. ALBERT maintained similar performance levels while improving efficiency with reduced parameters and faster training. Feature-based method achieved excellent results, offering high accuracy and computational efficiency.

Transductive SVMs leverage both labeled and unlabeled data effectively, excelled when labeled data is scarce, but they were computationally intensive. Non-Transductive SVMs provided robust and efficient classification using only labeled data, though they did not match the performance of deep learning models on complex datasets.

Overall, Feature-based, BERT and ALBERT lead in performance but at a higher computational cost, while SVMs and feature-based methods offered practical alternatives depending on resource constraints and dataset characteristics.

### 4.6 Scenario 4: Comparing Sub-Techniques within Traditional Methods of Artificial Intelligence Category

In this scenario, we evaluated and compared traditional artificial intelligence sub-techniques to assess their ongoing relevance and limitations in modern text classification tasks. The assessment was conducted using the Reuters-21578 dataset [10]. The experimental results are illustrated in Figure 9.

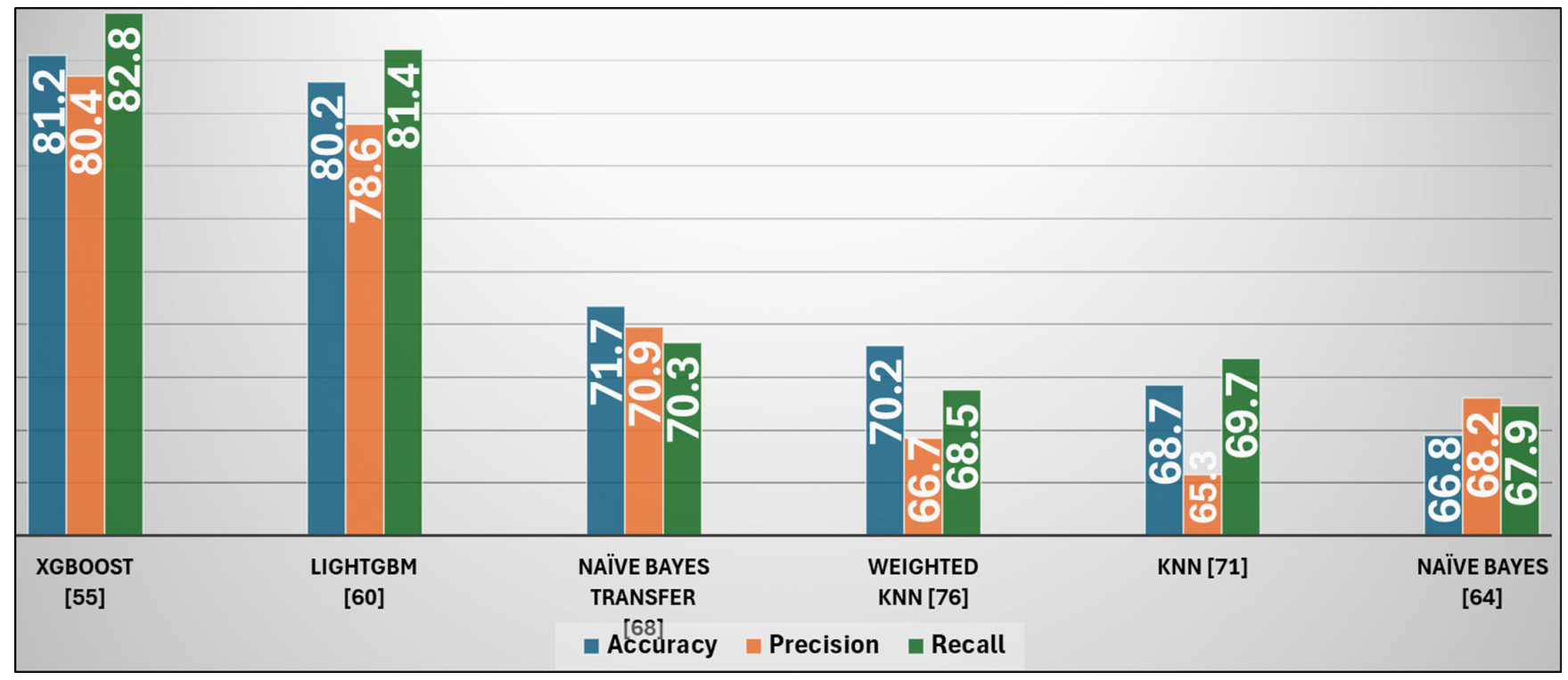

**Fig. 9:** Depicting the results of comparing Traditional Artificial Intelligence sub-techniques

Upon analyzing the experimental outcomes, it was observed that classical algorithms incorporating gradient-boosted decision trees (XGBoost and LightGBM), were highly effective in accurately classifying text and managing missing data, resulting in high accuracy scores. However, these algorithms tended to overfit small datasets. Their improved performance was mainly attributed to their ability to retain instances with large gradients while disregarding those with small gradients.

The Probabilistic Graphical Models (Naïve Bayes and Naïve Bayes Transfer) showed modest accuracy scores due to their assumption of feature non-interdependence. This assumption means the algorithms do not consider the relationships and interactions between different features of the dataset. Despite this limitation, these models achieved some success in estimating an initial model by utilizing labeled data from the distribution.

The incorporation of distance-based classification and normalization of training data in Nearest Neighbor-Based methods (KNN and Weighted KNN), proved beneficial, achieving satisfactory accuracy scores. However, these algorithms faced challenges due to the presence of irrelevant features and inconsistent feature scaling. Irrelevant features can introduce noise and bias, while inconsistent scaling can cause some features to dominate others due to their larger scale, leading to inaccuracies in classification.

### 4.7 Scenario 5: Comparing Sub-Techniques within Deep Learning Methods of Artificial Intelligence Category

In this scenario, we evaluated and compared deep learning artificial intelligence sub-techniques for their effectiveness in handling large-scale, diverse text data, particularly for sentiment analysis and product review classification. These methods were assessed using the Amazon product reviews dataset [11] to support their adoption in commercial applications. Figure 10 depicts the experimental results.

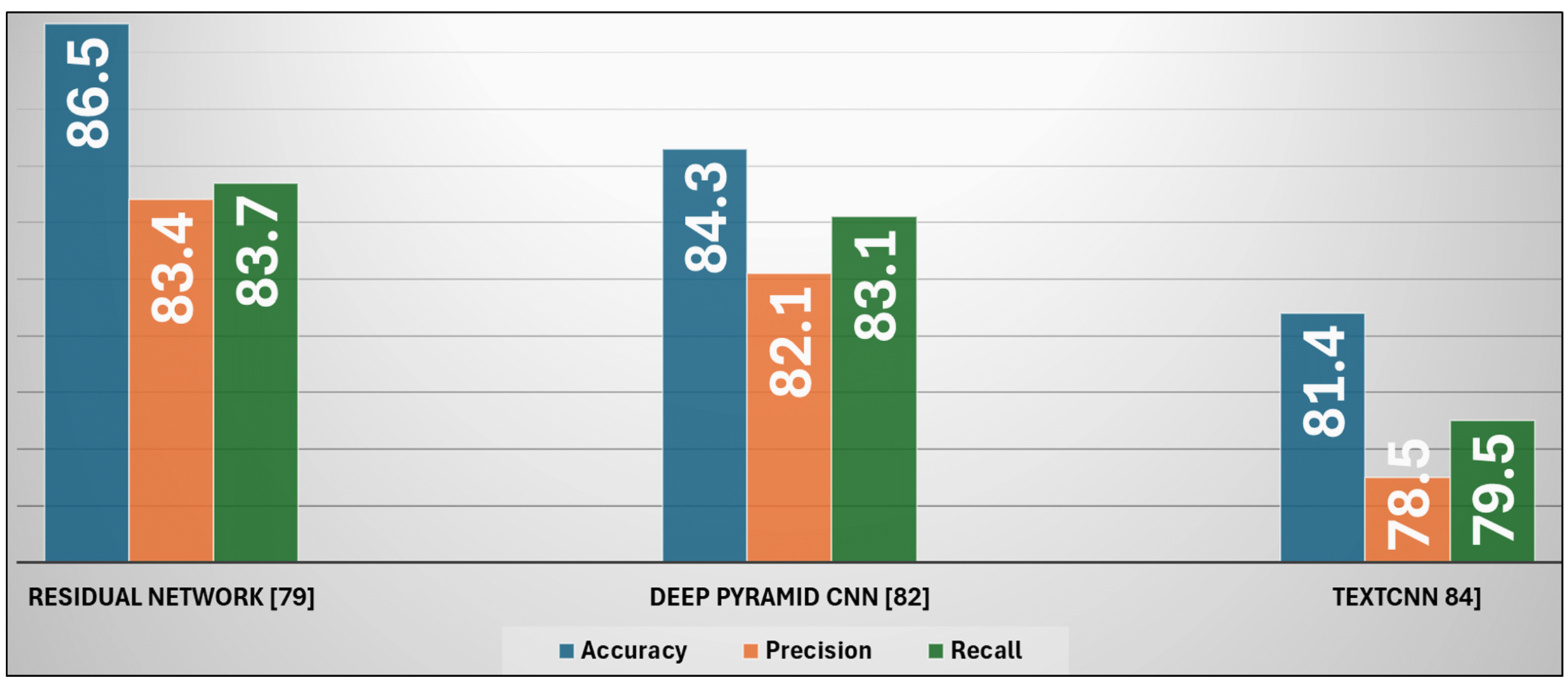

**Fig. 10:** Depicting the results of comparing Deep Learning Artificial Intelligence sub-techniques

ResNet, with its skip connections, effectively mitigated the vanishing gradient problem, leading to improved training efficiency and higher accuracy on complex datasets. Deep Pyramid CNN, designed for deep hierarchical feature extraction, excelled in capturing long-range dependencies and intricate text patterns, resulting in robust performance and high accuracy.

TextCNN, known for its simplicity and effectiveness, performed well by leveraging multiple convolutional filters to capture various n-gram features, achieved satisfactory accuracy with relatively low computational requirements. However, ResNet and Deep Pyramid CNN outperformed TextCNN in capturing deeper contextual information due to their more sophisticated architectures.

Overall, while TextCNN offered a balanced trade-off between performance and efficiency, ResNet and DPCNN provided superior accuracy and robustness, making them more suitable for complex and large-scale text classification tasks.

## 5 Challenges and Potential Future Perspectives for Text Classification

1. **Continual Learning**: Continual learning is a type of machine learning that enables models to learn from a continuous data stream. For text classification, it can aid in adapting to new language use patterns and user preferences, boosting accuracy and adaptability. This approach ensures that models remain relevant and effective over time, accommodating the dynamic nature of language and user interactions.

2. **Interpretable Models**: As text classification is increasingly used in high-stakes applications, interpretability is crucial for building trust. More research on interpretable models for text classification is expected to provide insights into prediction mechanisms. By making models more transparent, stakeholders can better understand and trust the decisions made by these systems, which is especially important in fields like healthcare, finance, and legal services.

3. **Deep Learning Models**: Deep learning models, such as CNNs, RNNs, and GNNs, are highly efficient in text classification but require optimization of technical constraints like layer depth, regularization, and network learning rate for optimal performance. A significant challenge lies in enhancing model robustness against adversarial samples, which can significantly reduce efficacy. Although deep neural networks (DNNs) excel in feature extraction and semantic mining, designing precise models for diverse applications requires a deeper understanding of underlying theories. Improving model performance and interpretability remains an ongoing challenge due to the lack of clear guidelines for optimization and the often unexplainable way in which deep learning models learn. As research advances, creating more robust and transparent deep learning models will be critical for their broader application and acceptance.

4. **Multilingual Text Classification**: Handling multilingual text data presents a significant challenge for text classification models. Developing models that can accurately classify text across different languages and dialects requires advanced natural language processing techniques and a deep understanding of linguistic nuances. Future research may focus on creating more sophisticated multilingual models that can seamlessly integrate and understand multiple languages, thereby improving their applicability in global contexts.

5. **Handling Imbalanced Datasets**: Imbalanced datasets, where some classes are underrepresented, pose a major challenge for text classification. Models trained on such datasets often exhibit bias towards the majority class, leading to poor performance on minority classes. Future work could explore advanced data augmentation techniques, resampling methods, and cost-sensitive learning strategies to better handle class imbalance. Additionally, developing evaluation metrics that accurately reflect model performance on imbalanced datasets will be crucial.

6. **Privacy and Ethical Considerations**: As text classification models increasingly handle sensitive and personal data, ensuring privacy and adhering to ethical standards is paramount. Future perspectives may involve developing privacy-preserving algorithms that can perform text classification without compromising user data. Techniques such as federated learning, differential privacy, and secure multi-party computation can help mitigate privacy concerns. Additionally, establishing ethical guidelines and frameworks for the responsible use of text classification technologies will be essential to prevent misuse and ensure fairness

# 6 Conclusion

This survey paper introduced a comprehensive taxonomy specifically designed for text classification based on research fields. The taxonomy is structured into hierarchical levels: research field-based category, research field-based sub-category, methodology technique, methodology sub-technique, and research field applications. By employing a dual evaluation approach—empirical and experimental—we provided a detailed and nuanced understanding of text classification algorithms and their applications. This empowers researchers to make informed decisions based on precise, field-specific insights.

In addition to proposing a detailed taxonomy, our study includes rigorous empirical evaluations. We conducted experimental evaluations under five scenarios, focusing on different aspects of text classification:

1. Recurrent Neural Network-Based Data Science Classification for sentiment and sequence classification tasks using the Yelp reviews dataset.
2. Embedding-Based Data Science Classification for managing text with diverse formats and errors using the BBC News dataset.
3. Pre-Trained-Based Classification Data Science for measuring performance in transfer learning for sentiment analysis using the SST-2 dataset.
4. Traditional Artificial Intelligence Techniques for assessing their relevance in contemporary text classification tasks using the Reuters-21578 dataset.
5. Deep Learning Artificial Intelligence Techniques for handling large-scale, diverse text data for sentiment analysis and product review classification using the Amazon product reviews dataset.

Our evaluations highlighted the subtle distinctions between closely related algorithms and techniques, aiding researchers in selecting the most appropriate methods for their specific tasks. For instance, in Scenario 1, recurrent neural network-based techniques showed some ability to acquire grammatical connections but struggled with complex sentence structures, while advanced inference techniques like LSTMs provided superior accuracy in capturing intricate language patterns. Scenario 2 revealed that Capsule Networks excelled in recognizing hierarchical relationships within text, enhancing their ability to manage complex patterns and errors, compared to Bi-LSTM.

In Scenario 3, the feature-based method excelled by extracting meaningful features such as term frequency, TF-IDF, and word embeddings, reducing dimensionality to improve efficiency and accuracy. BERT and ALBERT achieved high accuracy due to their contextual understanding, though BERT required substantial computational resources, while ALBERT improved efficiency with reduced parameters. While feature-based methods, BERT, and ALBERT led in performance but at higher computational costs, SVMs and feature-based methods offered practical alternatives depending on resource constraints and dataset characteristics.

Traditional AI techniques, as evaluated in Scenario 4, demonstrated the continued relevance of algorithms like gradient-boosted decision trees, although they tended to overfit some datasets. Finally, Scenario 5 showcased the effectiveness of deep learning techniques like ResNet and Deep Pyramid CNN in handling large-scale text data, outperforming simpler models like TextCNN in capturing deeper contextual information.

Through our research field-based taxonomy, empirical evaluations, and experimental comparisons, researchers can gain a nuanced and comprehensive understanding of text classification algorithms. This in-depth knowledge empowers them to make well-informed decisions, optimizing the impact and effectiveness of their research in the field of text data mining.